\PassOptionsToPackage{dvipsnames,table}{xcolor}
\documentclass{article}
\PassOptionsToPackage{numbers, compress}{natbib}
\usepackage[preprint]{neurips_2024}
  \usepackage[utf8]{inputenc}
\usepackage[T1]{fontenc}
\usepackage{url}
\usepackage{booktabs}
\usepackage{amsfonts} 
\usepackage{nicefrac}
\usepackage{microtype}
\usepackage{tcolorbox}
\tcbuselibrary{breakable, skins}
\usepackage{amsmath}
\usepackage[dvipsnames]{xcolor}
\usepackage{multirow}
\usepackage{colortbl}
\usepackage{graphicx}
\usepackage{adjustbox}
\usepackage{rotating}
\usepackage{array}
\usepackage{tabularx}
\usepackage{xltabular}
\usepackage{placeins}
\usepackage{float}
\usepackage{makecell}
\usepackage{subcaption}
\usepackage{tcolorbox}
\usepackage{longtable}
\usepackage{xurl}
\tcbset{
    tightbox/.style={
        colback=blue!5, 
        colframe=darkbluegrey, 
        boxsep=2pt,
        before skip=3pt,
        after skip=3pt,
        fontupper=\linespread{0.9}\selectfont
    }
}
\definecolor{rank1}{RGB}{198, 239, 206}
\definecolor{rank2}{RGB}{218, 238, 190}
\definecolor{rank3}{RGB}{238, 237, 174}
\definecolor{rank4}{RGB}{255, 235, 156}
\definecolor{rank5}{RGB}{255, 213, 141}
\definecolor{rank6}{RGB}{255, 191, 148}
\definecolor{rank7}{RGB}{255, 169, 164}

\definecolor{blacktitle}{rgb}{0.1, 0.1, 0.1}
\definecolor{darkbluegrey}{rgb}{0.2, 0.3, 0.5}

\newcommand{\Dist}[1]{\mathrm{Dist}_{#1}}
\newcolumntype{L}[1]{>{\raggedright\arraybackslash}m{#1}}

\title{On the Importance and Evaluation of Narrativity \\ in Natural Language AI Explanations}

\author{
  Mateusz Cedro\thanks{Corresponding author: mateusz.cedro@uantwerpen.be}
  \quad \quad
  David Martens \\
  University of Antwerp, Belgium
}
\date{}

\begin{document}
\maketitle

\setcounter{footnote}{0}

\begin{abstract}
Explainable AI (XAI) aims to make the behaviour of machine learning models interpretable, yet many explanation methods remain difficult to understand. The integration of Natural Language Generation into XAI aims to deliver explanations in textual form, making them more accessible to practitioners. Current approaches, however, largely yield static lists of feature importances. Although such explanations indicate \emph{what} influences the prediction, they do not explain \emph{why} the prediction occurs. In this study, we draw on insights from social sciences and linguistics, and argue that XAI explanations should be presented in the form of narratives. Narrative explanations support human understanding through four defining properties: continuous structure, cause-effect mechanisms, linguistic fluency, and lexical diversity. We show that standard Natural Language Processing (NLP) metrics based solely on token probability or word frequency fail to capture these properties and can be matched or exceeded by tautological text that conveys no explanatory content. To address this issue, we propose seven automatic metrics that quantify the narrative quality of explanations along the four identified dimensions. We benchmark current state-of-the-art explanation generation methods on six datasets and show that the proposed metrics separate descriptive from narrative explanations more reliably than standard NLP metrics. Finally, to further advance the field, we propose a set of problem-agnostic XAI Narrative generation rules for producing natural language XAI explanations, so that the resulting XAI Narratives exhibit stronger narrative properties and align with the findings from the linguistic and social science literature.
\end{abstract}

\section{Introduction}\label{sec:introduction}
The field of Explainable AI (XAI) aims to make the decision-making processes of complex machine learning models transparent and understandable~\cite{miller2019explanation}. However, the vast majority of developed methods produce highly technical explanations that remain inaccessible to non-experts and challenging even for practitioners, inadvertently undermining trust in AI systems~\cite{miller2019explanation,tellmeastory,miller2017explainable}. The integration of natural language generation into XAI promises to address that limitation by providing explanations in the most intuitive medium of communication~\cite{tellmeastory,dahlstrom2014using,graphxain,zytek2024explingo,talktomodel2023explaining,friedman1974explanation}. However, the mere translation of feature attributions into text does not guarantee understanding. Although current natural language methods can generate grammatically correct sentences, they often produce static, descriptive, and repetitive list-like accounts of feature importances that explain \emph{what} the model predicted, but fail to provide a narrative explanation of \emph{why} a specific prediction occurs.~\cite{tellmeastory,graphxain}. This distinction is essential, as research in social sciences and linguistics identifies the narrative as a primary mechanism for human reasoning and comprehension~\cite{friedman1974explanation,norris2005theoretical,bruner2009actual}. By overlooking the rhetorical structure of explanations, current XAI systems force users to cognitively reconstruct the ``story'' from disjointed facts, undermining the very transparency they aim to provide~\cite{miller2019explanation,tellmeastory}. 

Despite this cognitive foundation for narrative thinking, current XAI research does not systematically distinguish between \textit{descriptive} and \textit{narrative} explanations. In this paper, we address this issue by establishing the conceptual and practical differences between descriptive and narrative XAI explanations. Specifically, we:
\begin{itemize}
\item show that natural language explanations are not homogeneous, but can be classified as either descriptive or narrative depending on their structure and cognitive function,
\item identify four specific narrative properties (continuous structure, cause-effect mechanisms, linguistic fluency, and lexical diversity) grounded in social sciences that support human understanding in the XAI context,
\item demonstrate that standard natural language processing (NLP) automatic metrics, such as perplexity and word-frequency statistics, do not capture narrative quality,
\item develop seven metrics to quantify the \emph{narrativity} of natural language explanations,
\item formulate a set of rules based on the identified narrative properties to guide the generation of natural language narrative explanations, and
\item benchmark a range of XAI methods that generate natural language explanations using proposed metrics to estimate the extent to which their outputs exhibit narrative properties.
\end{itemize}

We argue that explicitly distinguishing descriptive from narrative explanations, providing tools to develop XAI Narratives, and quantify their narrativity, are necessary steps towards truly human-centred XAI.

\section{Background and Related Work}\label{sec:background}

The field of XAI has developed numerous post-hoc explanation techniques for tabular~\cite{lundberg2017unified,ribeiro2016should,martens2014explaining}, image~\cite{simonyan2013deep,Selvaraju2017Grad_cam}, text~\cite{lundberg2017unified,ribeiro2016should,martens2014explaining}, and graph machine learning models~\cite{ying2019gnnexplainer}. However, many resulting explanations remain too complex for non-expert comprehension, forcing users to expend significant cognitive effort interpreting them~\cite{miller2017explainable}. \citet{miller2017explainable} refer to this as ``\textit{inmates running the asylum}'', arguing that explanations should assign or develop meaning~\cite{norris2005theoretical} rather than merely listing feature importances or exposing internal computations. One response is to express explanations in natural language, with the goal of communicating model reasoning in forms that align better with how practitioners process new knowledge.

\subsection{Natural Language Explanations in XAI}

Recently, researchers have introduced methods to convert technical post-hoc outputs into natural language, offering a more accessible format~\cite{tellmeastory,cambria2023survey}. Some approaches rely on textual templates with placeholders instantiated by features, usually ordered by their estimated importance~\cite{talktomodel2023explaining,LI2018301}. Others use generative models, mainly Large Language Models (LLMs) to generate explanations, either in a zero-shot setting~\cite{tellmeastory,graphxain,pan2025graphnarrator,sammani2025zeroshot,he2026agentic} or using in-context few-shot learning~\cite{zytek2024explingo,zhou2023}.

While template-based explanations are controllable, their static structure often generates unnatural and rigid text~\cite{graphxain,mariotti2020towards}. On the other hand, although LLM-generated explanations offer greater flexibility and fluency, they remain less controllable and prone to hallucinations~\cite{mariotti2020towards}. Furthermore, the vast majority of existing approaches lead to \emph{XAI Descriptions}, defined as a static presentation of features relevant to a prediction, conveyed in a context-free, purely fact-based manner~\cite{graphxain}. Such purely descriptive forms are often insufficient in conveying the explanation to the user in the understandable way. Instead, explanations should ideally take the form of an \emph{XAI Narratives}, which is a structured story-like representation of a prediction that emphasizes the relationships between key features in a context-dependent manner~\cite{graphxain}. Narrative explanations enhance understanding by answering not only \textit{what} happened, but fundamentally \textit{why} it happened~\cite{tellmeastory}. Table~\ref{tab:desc_vs_narr} illustrates the distinction between descriptive and narrative explanations using a simplified credit risk example, and shows how feature importance scores can be verbalised either as a list of contributions or as a connected account of \emph{why} the prediction is made.

\begin{table}[t!]
\caption{Examples of explanation approaches for a high credit risk prediction: a feature attribution method (SHAP), an XAI Description, and an XAI Narrative.}
\label{tab:desc_vs_narr}
\centering
\begin{tabularx}{\textwidth}{X}
\toprule
\textbf{Explanation of Model Prediction} \\
\midrule
    \textbf{Feature attribution (SHAP)}:
    $\phi_{\text{credit amount}}: +0.13$ (\textit{3{,}632 EUR}), 
    $\phi_{\text{age}}: -0.12$ (\textit{22}), 
    \newline $\phi_{\text{housing}}: -0.04$ (\textit{rent}),
    $\phi_{\text{purpose}}: -0.03$ (\textit{car}),
    $\phi_{\text{sex}}: -0.03$ (\textit{female})\\
    \midrule
    \textbf{XAI Description}:  
    ``The model predicts Good Credit Risk. The most important feature is Credit amount (3,632 EUR), with a SHAP contribution of $+0.13$. The second most important feature is Age (22), with a SHAP contribution of $-0.12$. The third most important feature is Housing (rent), with a SHAP contribution of $-0.04$. The fourth most important feature is Purpose (car), with a SHAP contribution of $-0.03$. The fifth most important feature is Sex (female), with a SHAP contribution of $-0.03$.'' \\
    \midrule
    \textbf{XAI Narrative}:
    ``The model predicts Good Credit Risk for this applicant. It starts by looking at the requested credit amount of 3,632 EUR, which on its own pushes the decision toward bad credit risk. Against this starting point, the applicant's age of 22 acts as the strongest counterweight and pulls the assessment back toward good credit risk. Building on that shift, the applicant's status as a renter further softens the initial financial concern. In turn, the car loan purpose and the recorded female sex add smaller effects in the same favourable direction. As a result, these combined factors outweigh the earlier negative signal from the credit amount, and the model therefore classifies the applicant as a Good Credit Risk.'' \\
\bottomrule
\end{tabularx}
\end{table}

\subsection{Evaluating Quality of Natural Language Text}
\paragraph{Perplexity and Statistical Metrics.}
Perplexity (PPL)~\cite{jelinek1977perplexity} is frequently used as a reference-free proxy for linguistic quality and plausibility~\cite{ichmoukhamedov2024good,shirvani2025rule2text}. However, it is an unreliable stand-alone indicator of text quality~\cite{wang2022perplexity,hashimoto2019unifying}, since it is sensitive to text length, can be artificially lowered by repetition, and does not strictly correlate with human perception of text quality~\cite{hashimoto2019unifying,chaganty2018price,velivckovic2026perplexity,fang2024wrong}. To address this disconnect between perplexity and quality, their proposed framework combines model probabilities with human judgments. Nevertheless, this approach cannot be computed automatically, which limits its usability.

Metrics such as BLEU~\cite{papineni2002bleu}, METEOR~\cite{banerjee2005meteor}, ROUGE~\cite{lin2004rouge}, and BLEURT~\cite{sellam2020bleurt} score a generated text by comparing it with one or more reference texts. BLEU, METEOR, and ROUGE rely mainly on lexical overlap and are known to correlate only weakly with human judgments in many generation settings. BLEURT instead uses a learned scorer based on BERT, trained to better match human ratings while still operating on candidate–reference pairs. However, in XAI, reference-based metrics are often inapplicable since the reference explanations are rarely available.

\citet{meister2021language} propose evaluating text by assessing adherence to statistical laws of natural language, such as Zipf's law (rank-frequency) and Heaps' law (type-token). While this provides insight into how closely models mimic the macro-level structural properties of human language, it remains a measure of distributional similarity rather than inherent generation quality or semantic coherence.

\FloatBarrier

\begin{table}[!htbp]
\caption{Overview of existing evaluation methods of the quality of the text.}
\centering
\small
\renewcommand{\arraystretch}{1.1}
\setlength{\tabcolsep}{4pt}
\begin{tabularx}{\textwidth}{L{1.7cm} L{2.6cm} X X}
\toprule
Type & Article & Focus (Metric/Framework) & Description \\
\midrule
% QUANTITATIVE SECTION
\multirow[c]{35}{*}{Quantitative}
 & \multirow[c]{2}{*}{~\citet{zytek2024explingo}}
 & Accuracy, Completeness, Fluency, Conciseness
 & \emph{LLM-as-a-Judge} approach to evaluate generated explanations. \\
 \cmidrule(l){2-4}
 & \multirow[c]{5}{*}{\citet{papineni2002bleu}} 
 & \multirow[c]{5}{*}{Text Similarity (BLEU)}
 & Evaluates quality by comparing a machine-generated text against (human) reference translations, scoring similarity based on matching n-grams (word sequences). \\
  \cmidrule(l){2-4}
 & \multirow[c]{3}{*}{\citet{lin2004rouge}}
 & \multirow[c]{3}{*}{Text Similarity (ROUGE)}
 & Compares generated text against human reference summaries using recall-oriented overlap scores. \\
 \cmidrule(l){2-4}
 & \multirow[c]{3}{*}{\makecell[l]{Banerjee\\and Lavie~\cite{banerjee2005meteor}}}
 & \multirow[c]{3}{*}{Text Similarity (METEOR)}
 & Evaluates translations via generalized unigram matching against human references. \\
  \cmidrule(l){2-4}
 & \multirow[c]{3}{*}{\citet{sellam2020bleurt}} 
 & \multirow[c]{3}{*}{Semantic Similarity (BLEURT)}
 & A BERT-based~\cite{devlin2019bert} model evaluating text quality via semantic similarity to match human judgment. \\
 \cmidrule(l){2-4}
& \multirow[c]{3}{*}{\makecell[l]{Tekkesinoglu\\and Kunz~\cite{tekkesinoglu2024feature}}}
& \multirow[c]{3}{*}{\makecell[l]{Sociability, Causality,\\Selectiveness, Contrastiveness}}
& Assesses explanations across social dimensions by counting predefined cue words. \\
\cmidrule(l){2-4}
& \multirow[c]{4}{*}{\citet{jelinek1977perplexity}}
& \multirow[c]{4}{*}{Text Predictability (Perplexity)}
& Measures predictive uncertainty over a text sequence. Commonly used as a proxy for text fluency (token-to-token prediction). \\
 \cmidrule(l){2-4}
& \multirow[c]{4}{*}{\makecell[l]{Meister\\and Cotterell~\cite{meister2021language}}}
& \multirow[c]{4}{*}{Statistical Tendencies}
& Assesses the linguistic fit of generated text by measuring its adherence to the statistical regularities of natural language. \\
 \cmidrule(l){2-4}
& \multirow[c]{3}{*}{\citet{graesser2004coh}}
& \multirow[c]{3}{*}{Discourse Cohesion (Coh-Metrix)}
& Quantify discourse cohesion through lexicon- and feature-based indices in text. \\
 \cmidrule(l){2-4}
& \multirow[c]{4}{*}{\citet{sap2020recollection}}
& \multirow[c]{4}{*}{\makecell[l]{Sequence Likelihood\\(Narrative flow)}}
& Compares unordered (``bag'') and sequential (``chain'') text models by measuring how well each predicts narrative progression. \\
\midrule
% QUALITATIVE SECTION
\multirow[c]{10}{*}{Qualitative} 
 & \multirow[c]{3}{*}{\citet{tellmeastory}}
 & Convincingness, Ease, Confidence, Speed of understanding, Usability
 & Conduct user surveys to compare narrative explanations against the listed feature importances. \\
 \cmidrule(l){2-4}
 & \multirow[c]{4}{*}{\makecell[l]{Cedro\\and Martens~\cite{graphxain}}}
 & Understandability, Trustworthiness, Insightfulness, Satisfaction, Confidence, Convincingness, Communicability, Usability
 & Evaluate the clarity and usability of narrative explanations specifically for Graph Neural Network predictions. \\ 
 \cmidrule(l){2-4}
 & \multirow[c]{3}{*}{\citet{talktomodel2023explaining}}
 & \multirow[c]{3}{*}{\makecell[l]{Understandability, Confidence,\\Speed of understanding, Usability}}
 & Evaluate the user satisfaction among ML practitioners and lay users. \\ 
 \midrule
 % HYBRID
 \multirow[c]{9}{*}{Hybrid}
 & \multirow[c]{4}{*}{\makecell[l]{Ichmoukhamedov\\et al.~\cite{ichmoukhamedov2024good}}}
 & Faithfulness, Assumptions, Human Similarity (Rank Agreement, Sign Agreement, Value Agreement)
 & Measures the fidelity of generated explanations with respect to the original feature attribution scores and human text reference. \\
 \cmidrule(l){2-4}
 & \multirow[c]{5}{*}{\citet{hashimoto2019unifying}} 
 & \multirow[c]{5}{*}{Diversity, Quality (HUSE)}
 & Combines model probability with human judgment to evaluate diversity and quality, based on the error rate to predict if a sentence is human- or machine-generated. \\
 \bottomrule
\end{tabularx}
\label{tab:natural_language_methods}
\end{table}

\paragraph{Discourse and Readability Measures.}
Traditional readability formulas (e.g., Flesch-Kincaid or Gunning Fog Index) estimate text difficulty using linguistic features such as sentence length and word complexity, but fail to differentiate between text genres~\cite{graesser_computationalanalyses}. Coh-Metrix~\cite{graesser2004coh} addresses this by analysing cohesion and coherence through lexical, syntactic, and discourse-based features, yet it quantifies these dimensions via restricted lexicons. Relying solely on word frequency or lexicon matching is insufficient to assess text quality~\cite{wang2022perplexity}, as it fails to capture the structural dependencies and narrative flow essential for better understanding. Consequently, these tools assess general text features but are not designed to distinguish between expository, list-like descriptions and desirable narrative forms.

\paragraph{Narrativity.}
\citet{sap2020recollection} propose a \emph{narrative flow} metric to assess contextual coherence by comparing the negative log‑likelihood of a sentence under two language‑model configurations: one that scores the sentence in isolation and one that scores it when conditioned on the preceding text. A defining characteristic of their setup is that both the independent and the context‑conditioned models additionally condition on a provided ``theme'' or summary of the story. Although the approach is appropriate for their goal of distinguishing imagined from recalled stories, it cannot be used to assess the XAI explanations. The generation of post-hoc explanations aims to produce novel text describing model behaviour, meaning no ground-truth summary is available to condition the language models. The requirement for a pre-defined theme therefore makes the metric unsuitable for this domain.

\paragraph{Evaluation of Natural Language Explanations.} 
Assessing the quality of generated natural language explanations is not trivial~\cite{electronics10050593,ichmoukhamedov2024good}. Most frameworks rely on qualitative user studies~\cite{tellmeastory,graphxain,talktomodel2023explaining}. Nevertheless, user studies are resource-intensive and often difficult to reproduce across datasets and tasks, which limits their use for systematic benchmarking.

Recent automated frameworks have attempted to address this limitation. \citet{ichmoukhamedov2024good} evaluate the fidelity of LLM-generated explanations with respect to original feature attribution scores. Although valid for checking faithfulness, this framework does not assess whether the explanation is an XAI Description or an XAI Narrative. Similarly, \citet{tekkesinoglu2024feature} evaluate LLM-generated explanations along dimensions identified by \citet{miller2019explanation}: sociability, causality, selectiveness, and contrastiveness. Their implementation is primarily lexicon‑based, meaning that sociability is measured by counting the frequency of greeting terms (e.g., \emph{hello}). As a result, the findings capture the presence of cue words rather than an underlying explanatory structure. Table~\ref{tab:natural_language_methods} summarizes selected frameworks for evaluating natural language text.

\section{Narrative vs.~Descriptive Explanations: Insights from Social Sciences and a Conceptual Framework}
\label{sec:descriptive_vs_narrative}

Although the distinction between narrative and descriptive text is well-established in linguistics and social sciences~\cite{dahlstrom2014using,friedman1974explanation,salmon1984scientific}, its systematic application to XAI remains underexplored. Our work can be seen as a continuation of the research direction that aims to make AI explanations understandable for a broader audience, as emphasized by Miller~\cite{miller2019explanation,miller2017explainable}, who argues that explanations should be grounded in research from the social sciences. Supporting this interdisciplinary approach, \citet{norris2005theoretical} observe that ``\textit{Narrative explanations naturally must take on features both of narratives and of explanations.}'' In the following sections, we build on this work by translating qualitative insights into measurable features, to define and evaluate narrative explanations for model interpretability. In Table~\ref{tab:lit_review_metrics} we summarise the theoretical concepts and the proposed evaluation metrics that we introduce throughout this section.

\subsection{Continuous Structure of Narrative Explanation}
\label{subsec:features_continuous_structure}

Narrative explanations rely on continuity to transform a sequence of events into a structurally coherent text. Unlike a temporal listing of discrete events or a set of independent facts, a narrative links observations into a continuous structure in which meaning depends on the specific ordering of the sequence~\cite{miller2019explanation,norris2005theoretical,salmon1989four}, that is, ``\textit{the story leading up to the event to be explained}''~\cite{salmon1989four}. Narrative form further implies an organised structure, typically with a beginning, middle, and end~\cite{norris2005theoretical}. Deictic Shift Theory suggests that unmarked discontinuities disrupt readers' situation models, increasing the inferential effort required to integrate disparate elements of the text~\cite{murray1997connectives,graesser2011coh,galbraith1995deictic}. Structurally, this corresponds to ``\textit{interweaving sequential but overlapping threads}'' so that the text functions as a relatively independent whole rather than a collection of isolated data points~\cite{norris2005theoretical}. 

Moreover, as noted by~\citet{norris2005theoretical}, a narrative requires that events be connected so that ``\textit{individual events can be seen in the perspective of the others}''. To preserve the continuous structure, narrative explanations depend on connectives to serve as explicit linguistic markers that link elements of the narrative together. \citet{graesser2011coh} show that connectives form an independent category in narrative texts because they establish the structural cohesion of the situation model.

\paragraph{Principle.}
We emphasise the principle of \emph{continuous structure} as stated by~\citet{murray1997connectives}: ``\textit{The relevance of this perspective to the present research is the principle of continuity, which lies at the heart of deictic shift theory. (...) When a reader encodes a text event that is discontinuous in the absence of a marker or indication of the discontinuity, reading is more difficult.}'' To prevent such structural discontinuity, narratives rely on connectives to serve as explicit linguistic markers. \citet{graesser2011coh} detail this specific function, stating that: ``\emph{Connectives also constitute an important category \emph{[for the narrative text]} because they are important in establishing cohesion of the situation model.}''

\newpage
\paragraph{Implementation in XAI.}
We emphasize the following rule:

\begin{tcolorbox}[tightbox, title=\textbf{The continuous structure rule}]
\setlength{\parskip}{5pt}
An XAI Narrative should establish a continuous structure by following a clear narrative arc with a beginning, middle, and end, while using explicit linguistic connectives so that individual events can be seen in the perspective of the others.
\end{tcolorbox}

\begin{table}[!t]
\centering
\caption{Narrative concepts grounded in social sciences and linguistics literature. For each concept, supported by relevant citations, we propose metrics to measure it in XAI explanations (arrows indicate whether higher ($\uparrow$) or lower ($\downarrow$) values denote a stronger narrative form).}
\small
\begin{tabularx}{\textwidth}{p{1.35cm} p{1.0cm} X p{3.9cm}}
\toprule
Principle & Literature & Quote Example & Metric (Narrative Direction) \\
\midrule
\multirow{6}{*}{\makecell[l]{Continuous\\Structure}}
  & \multirow{6}{=}{\cite{salmon1989four,murray1997connectives,o1998updating,graesser2011coh}}
  & ``\emph{(...) cohesion breaks result in an increase in reading time and generation of inferences.}''~\cite{graesser2011coh}
  & \makecell[l]{\emph{Continuous Structure Rate} \\ (CSR) $\uparrow$, \\ \emph{Diversity-adjusted Context} \\ \emph{Progression Rate} (DCPR) $\downarrow$} \\
\cmidrule{3-4}
  & & ``\emph{Connectives also constitute an important category} [for the narrative text] \emph{because they are important in establishing cohesion of the situation model.}''~\cite{graesser2011coh}
  & \multirow{4}{*}{\makecell[l]{\emph{Connectives-adjusted Context} \\ \emph{Progression Rate} (CCPR) $\downarrow$}} \\
\midrule
\multirow{8}{*}{\makecell[l]{Cause-Effect\\Mechanisms}}
  & \multirow{8}{=}{\cite{norris2005theoretical,simon2000discovering,salmon1989four,hempel1948studies}}
  & ``\emph{(...) there is a narrative or genetic type of explanation that consists in telling the story leading up to the event to be explained. Since the mere recital of just any set of preceding occurrences may have no explanatory value whatever, the narrative must involve events that are causally relevant to the explanandum} [what is being explained] \emph{if it is to serve as an explanation}''~\cite{salmon1989four} & \multirow{8}{*}{\makecell[l]{
  % \emph{Cause-Effect Ratio} (CER) $\uparrow$, \\
  \emph{Cause-Effect-adjusted Context} \\ \emph{Progression Rate} (CECPR) $\downarrow$ }} \\
\midrule
\multirow{3}{*}{\makecell[l]{Linguistic\\Fluency}}
  & \multirow{3}{*}{\cite{kellogg1991relative}}
  & ``\emph{(...) the habitual deployment of a narrative schema should support highly efficient or fluent narrative composition.}''~\cite{kellogg1991relative} & \multirow{3}{*}{\emph{Fluency-Diversity Rate} (FDR) $\uparrow$} \\
\midrule
\multirow{5}{*}{\makecell[l]{Lexical\\Diversity}}
  & \multirow{5}{*}{\cite{saenz2002examining,graesser2011coh}}
  & ``\emph{Lexical diversity is related to cohesion because a higher number of different words need to be integrated into the discourse context.}''~\cite{graesser2011coh}
  & \multirow{3}{*}{\makecell[l]{\emph{Type-Token-adjusted Context} \\ \emph{Progression Rate} (TTCPR) $\downarrow$}} \\
\cmidrule{3-4}
  & & ``\emph{(...) verbs are more prevalent in narrative discourse than in informational texts (...)}''~\cite{graesser2011coh}
  & \multirow{2}{*}{\makecell[l]{\emph{Verb-adjusted Context} \\ \emph{Progression Rate} (VCPR) $\downarrow$}} \\
\bottomrule
\end{tabularx}
\label{tab:lit_review_metrics}
\end{table}

\paragraph{Measurement.}
We propose two metrics that quantify the principle of \emph{continuity} to distinguish narrative from descriptive explanations. First, \textit{Continuous Structure Rate} (CSR) ($\uparrow$) measures the sensitivity of perplexity (PPL) to sentence-order perturbations, relative to the perplexity of originally ordered text. We compute the CSR as:
\begin{equation}\label{eq:csr}
    \text{Continuous Structure Rate} = \frac{\mathrm{PPL}_{\text{shuffled}} - \mathrm{PPL}_{\text{ordered}}}{\mathrm{PPL}_{\text{ordered}}}
\end{equation}
Sentence reordering in a narrative explanation is expected to increase the perplexity considerably, since it disrupts the continuous structure of the text. In contrast, reordering sentences in a descriptive, list-like explanation should have a smaller effect, because expository descriptions typically present loosely related facts without strong inter-sentential dependence~\cite{norris2005theoretical,weaver1991expository}. Using the relative change in perplexity under shuffling as a continuity measure also mitigates two limitations of perplexity noted by~\citet{wang2022perplexity}, namely sensitivity to text length and sensitivity to repetition, as shuffling preserves both the number of tokens and the amount of repeated content. This property is not captured by prior evaluation frameworks such as \citet{shirvani2025rule2text} and \citet{sap2020recollection}.

\paragraph{\emph{Continuous Structure Rate} Example.}
For the example in Table~\ref{tab:desc_vs_narr}, we first compute the perplexity of each explanation in its original sentence order, obtaining $\text{PPL}_{\text{ordered}} = 15.10$ for the XAI Narrative and $\text{PPL}_{\text{ordered}} = 4.65$ for the XAI Description. We then randomly shuffle the sentences (seed $= 42$) and recompute the perplexity, yielding $\text{PPL}_{\text{shuffled}} = 22.89$ and $6.49$, respectively. Substituting these values into Equation~\eqref{eq:csr}, the XAI Narrative attains a CSR ($\uparrow$) of $(22.89 - 15.10)/15.10 = 0.52$, while the XAI Description reaches $(6.49 - 4.65)/4.65 = 0.40$. The larger relative increase under shuffling for the XAI Narrative indicates that its sentence order carries meaningful structural information, whereas the smaller change for the XAI Description reflects weaker inter-sentential dependence, consistent with its list-like character.\footnote{Perplexities and relative changes are computed with \texttt{Llama-3.1-8B}. We share the results concerning examples from Table~\ref{tab:desc_vs_narr} at: \url{https://github.com/ADMAntwerp/On-the-Importance-and-Evaluation-of-Narrativity-in-Natural-Language-AI-Explanations/blob/main/tutorial/demo.py}}

Second, we evaluate the continuous structure of the text by introducing the \emph{Diversity-adjusted Context Progression Rate} (DCPR $\downarrow$). A continuous narrative interweaves sequential threads, meaning the accumulating context makes subsequent sentences more predictable. DCPR measures this context progression rate. The formulation ensures the metric penalizes artificial predictability gains caused by degenerate repetition rather than genuine narrative continuous structure.

We compute DCPR in three steps. First, we compute the set of cumulative perplexities for incremental text sequences $S[1:x]$. This process iteratively measures the perplexity after adding each subsequent sentence of the text. The calculation starts with the isolated first sentence ($x=1$) and finishes at the last sentence of the text, where the cumulative value equals the perplexity of the complete text. Second, motivated by the observation that predictability improves with context but approaches a non-zero asymptote determined by the text's intrinsic uncertainty~\cite{shannon1951prediction}, we fit a shifted exponential curve on the set of calculated cumulative perplexities:
\begin{equation}
\label{eq:exp_fit}
y(x) = A \cdot r^x + C
\end{equation}
where $y(x)$ denotes the cumulative perplexity value at step $x$. The parameter $C$ is the asymptotic baseline, which captures the irreducible uncertainty of the text domain, while $A$ quantifies the part of the uncertainty that is reduced as context accumulates. The parameter $r \in (0,1)$ controls the rate of contextual adaptation. In practice, we fit the equivalent form $y(x) = A e^{-bx} + C$ and map the decay constant to $r$ via $r = e^{-b}$. To ensure the metric is invariant to text length, we constrain the baseline $C$ to be upper-bounded by the minimum observed perplexity. This ensures that shorter narratives, which may share identical qualitative properties with longer ones, are evaluated against a realistic asymptote\footnote{A ``realistic asymptote'' corresponds to the information-theoretic lower bound of the specific text domain. Since natural language is non-deterministic, cumulative perplexity cannot decay to zero. For short sequences where the empirical plateau is not yet fully visible, we constrain the asymptote to the minimum observed value to prevent the model from overfitting a trajectory toward a theoretically impossible zero-perplexity state.}, preventing the bias where truncated sequences appear to decay faster simply because the stable tail is unobserved.

A lower value of $r$ indicates faster progression, that is, a more rapid decrease in perplexity as the text moves from its initial level ($A + C$) toward the asymptotic baseline ($C$). However, a small $r$ can also arise from degenerate repetition, since repeated phrases can reduce perplexity without adding coherent structure~\cite{holtzman2019curious,wang2022perplexity}. To reduce this effect, we weight the decay rate by the squared distinct bigram score $(\mathrm{Dist}_2)^2$, consistent with the non-linear repetition penalty used in FDR. We define the Diversity-adjusted Context Progression Rate as:
\begin{equation}\label{eq:dcpr}
\text{Diversity-adjusted Context Progression Rate} = \frac{r}{(\mathrm{Dist}_2)^2}
\end{equation}
This formulation separates predictability gains driven by a continuous narrative structure from those caused by degenerate looping. Introducing the asymptote $C$ reduces sensitivity to domain difficulty by disentangling the context progression rate $r$ from the baseline uncertainty captured by $C$. For narratives with a continuous structure, contextual dependencies yield fast convergence, producing a lower $r$, while maintaining high lexical diversity through a higher $(\mathrm{Dist}_2)^2$).This combination results in a lower DCPR. In contrast, descriptive lists achieve a low $r$ simply through repetition. Their diversity sharply decreases under these conditions, which leads to a penalized and higher DCPR. We therefore expect $\text{DCPR}_{\mathrm{Narrative}} < \text{DCPR}_{\mathrm{Descriptive}}$.

\paragraph{\emph{Diversity-adjusted Context Progression Rate} example.}
To compute the DCPR on the examples in Table~\ref{tab:desc_vs_narr}, we first obtain the cumulative perplexity as each sentence is appended. The XAI Narrative follows the trajectory $[117.86,\ 30.26,\ 17.79,\ 18.46,\ 20.97,\ 15.10]$, while the XAI Description follows $[491.33,\ 65.44,\ 17.75,\ 9.27,\ 6.27,\ 4.65]$. Fitting the shifted exponential in Equation~\eqref{eq:exp_fit} yields decay constants $r = 0.15$ for the XAI Narrative and $r = 0.13$ for the XAI Description. Although the description exhibits a marginally faster raw decay, its lower lexical diversity ($\text{Dist}_2 = 0.62$, compared with $0.92$ for the narrative) penalises the metric through the squared denominator in Equation~\eqref{eq:dcpr}. The XAI Narrative therefore achieves a DCPR ($\downarrow$) of $0.15 / 0.92^2 = 0.18$, whereas the XAI Description results in a score of $0.13 / 0.62^2 = 0.33$. This contrast confirms that the faster convergence observed in the description stems from repetition rather than from genuine continuous structure.

Third, we analyse the frequency of discourse connectives in each explanation. We focus on connectives because empirical work reports that explicit markers can reduce reading time, improve memory for clauses, and increase comprehension accuracy~\cite{murray1997connectives}, which are relevant for XAI explanations.

We detect connectives using the Lexicon of English Discourse Connectives~\citet{das2018constructing}. The lexicon contains 142 unique connective types and groups them into four relation classes. We define \emph{Connectives Ratio} (CR $\uparrow$) as the ratio of the number of connectives in the explanation to the total number of words. However, the explanation consisting of only connectives will result in gibberish text, yet resulting in $CR=1.0$, therefore, using CR is not enough to differentiate between the XAI Narratives and XAI Descriptions. Therefore, similarly to DCPR, we extend the CR metric to \emph{Connectives-adjusted Context Progression Rate} (CCPR $\downarrow$), which is the connectives weighting of the decay rate $r$ of text's perplexity. We define the CCPR as:
\begin{equation}\label{eq:ccpr}
\text{Connectives-adjusted Context Progression Rate} = \frac{r}{\mathrm{CR}^2}
\end{equation}
CCPR is expected to result in lower values for XAI Narratives than XAI Descriptions, as they consist of a higher number of connectives. We square the CR consistent with the non-linear repetition penalty used in the previous metric.

\paragraph{\emph{Connectives-adjusted Context Progression Rate} example.}
Reusing the decay constants from the shifted exponential curve fit ($r = 0.15$ for the XAI Narrative and $r = 0.13$ for the XAI Description), we weigh them by the squared connectives ratios, which we compute using the Lexicon of English Discourse Connectives~\cite{das2018constructing}. The XAI Narrative exhibits a higher density of discourse connectives ($\text{CR} = 0.096$) than the XAI Description ($\text{CR} = 0.071$). Substituting into Equation~\eqref{eq:ccpr}, the CCPR ($\downarrow$) for the XAI Narrative in Table~\ref{tab:desc_vs_narr} is $0.15 / 0.096^2 = 16.55$, whereas the XAI Description scores $0.13 / 0.071^2 = 25.10$. The lower CCPR of the narrative reflects both its more rapid contextual adaptation and its stronger reliance on connectives to establish cohesion between sentences, as emphasised by Graesser et al.~\cite{graesser2011coh}.

\subsection{Cause-Effect Mechanisms in Narrative Explanations}
\label{subsec:causal_dependencies_of_explanations}

Drawing on the philosophy of science, \citet{valentino2022scientific,valentino2024nature} argue that the primary function of an explanation is unification rather than mere deductive validity. This unification process situates the \textit{explanandum} within broader structural patterns~\cite{friedman1974explanation,kitcher1981,hempel1965aspects,salmon1984scientific}. To achieve this, an explanation must explicitly identify underlying cause-and-effect mechanisms. Consequently, evaluating natural language explanations requires measuring the presence of these causal structures instead of focusing solely on attribution faithfulness or fluency. This distinction establishes a clear boundary between descriptive and narrative explanations, as descriptive state only \textit{what} model predicted, whereas narrative address \textit{why} a prediction occurs by emphasizing the relevant cause-effect mechanisms~\cite{simon2000discovering,salmon1989four,hempel1948studies}.

\paragraph{Principle.} We emphasise the principle of \emph{cause-effect mechanisms} as proposed by~\citet{salmon1989four}: ``\emph{(...) there is a narrative or genetic type of explanation that consists in telling the story leading up to the event to be explained. Since the mere recital of just any set of preceding occurrences may have no explanatory value whatever, the narrative must involve events that are causally relevant to the explanandum \emph{[what is being explained]} if it is to serve as an explanation}''.

This distinction is important for cognitive processing and memory. \citet{simon2000discovering} recognizes a difference between the descriptions and explanations: ``\textit{Descriptions are often distinguished from explanations by saying that the former answer the question of how something behaves, the latter the question of why it behaves in this particular way}'' therefore, descriptions cannot be seen as a specific type of explanation, rather, as an entity that is detached from explanations.

\paragraph{Implementation in XAI.}
We emphasize the following rule: 
\begin{tcolorbox}[tightbox, title=\textbf{The cause-effect rule}]
\setlength{\parskip}{5pt}
An XAI Narrative should explicitly identify the underlying cause-effect mechanisms to clarify why the system made a particular prediction.
\end{tcolorbox}

\paragraph{Measurement.}
We quantify this dimension by measuring the frequency of explicit causal markers in the explanation. We define the \emph{Cause-Effect Ratio} (CER $\uparrow$) as the proportion of cause-effect markers relative to the total number of words. To compute this ratio, we compile a set of 19 specific cause-effect markers, including phrases such as ``as a result'', ``due to'', and `therefore''. These markers are distilled from the Lexicon of English Discourse Connectives~\cite{das2018constructing} and the Penn Discourse Treebank 3~\cite{prasad2019penn} (the complete list of these markers appears in the Appendix~\ref{app:cause_effect_lex}). However, relying solely on this ratio is insufficient evaluation, as the text composed only from the cause-effect markers will maximize the CER, but will result in gibberish text. Therefore, we extend the metric to the \emph{Cause-Effect adjusted Context Progression Rate} (CECPR $\downarrow$). This formulation applies the cause-effect ratio as an adjustment to the context progression rate $r$. We define the metric as:
\begin{equation}\label{eq:cecpr}
\text{Cause-Effect-adjusted Context Progression Rate} = \frac{r}{\mathrm{CER}^2}
\end{equation}
Narrative explanations structurally depend on causal markers to explain prediction mechanisms, so we expect them to achieve lower CECPR values than descriptive texts. We square the CER value to non-linearly penalize explanations that omit cause-effect relations. This formulation maintains consistency with the mathematical penalty applied in the Diversity-adjusted Context Progression Rate.

\paragraph{\emph{Cause-Effect-adjusted Context Progression Rate} example.}
Applying the proposed CECPR ($\downarrow$) metric to the examples in Table~\ref{tab:desc_vs_narr}, the XAI Narrative contains a single cause-effect marker (\textit{``as a result''}), giving a cause-effect ratio of $\text{CER} = 0.026$. Combined with the decay constant $r = 0.15$ obtained from Equation~\eqref{eq:exp_fit}, Equation~\eqref{eq:cecpr} yields $\text{CECPR} = 0.15 / 0.026^2 = 222.57$. The XAI Description, by contrast, contains no explicit causal markers ($\text{CER} = 0$), which renders the metric undefined due to division by zero. This behaviour is not an artefact of the formulation but a structural signal: the absence of causal markers reflects the absence of a cause-effect account in purely descriptive text, consistent with the distinction between descriptions and explanations articulated by Simon~\cite{simon2000discovering} and Salmon~\cite{salmon1989four}.

\subsection{Linguistic Fluency of Narrative Explanations}
\label{subsec:fluency}

Although an explanation structured with a beginning, middle, and end, as outlined in Section~\ref{subsec:features_continuous_structure}, provides necessary organization, it may still produce a rigid descriptive format rather than a fluent narrative. For instance, technical manuals typically follow a logical sequence but often depend on repetitive, list-like phrasing~\cite{norris2005theoretical,saenz2002examining}. Linguistic fluency addresses this limitation by presenting complex information in a more natural, accessible manner~\cite{kellogg1991relative}. Consequently, a structurally integrated and fluent narrative proves easier to read and comprehend than purely expository texts~\cite{graesser_computationalanalyses}, enabling the reader to construct an accurate mental model of the underlying system more readily~\cite{murray1997connectives}.

\paragraph{Principle.}
We emphasise the principle of fluency as proposed by~\citet{kellogg1991relative}, ``\emph{The habitual deployment of a narrative schema should support highly efficient or fluent narrative composition.}''

\paragraph{Implementation in XAI.}
The XAI Narratives should therefore help bridge the cognitive gap that arises when users are presented only with descriptive lists of features of post-hoc explanations by using natural language to organise discrete events into an intelligible, story-like narrative explanation. We emphasize the following rule:

\begin{tcolorbox}[tightbox, title=\textbf{The linguistic fluency rule}]
\setlength{\parskip}{5pt}
An XAI Narrative should explain the model's prediction with linguistic fluency, avoiding
repetitive, list-like structures.
\end{tcolorbox}

\paragraph{Measurement.} Perplexity~\cite{jelinek1977perplexity} is often used as an automatic metric of text quality, treating lower perplexity as an indicator of higher fluency~\cite{grave2016improving,lee2022_factualityenhanced}. However, previous work criticises perplexity as a standalone measure due to several limitations (e.g., repetitive text can achieve low perplexity despite being uninformative or hard to read, and the metric can be considerably affected by punctuation marks or text length)~\cite{wang2022perplexity,graesser2011coh,holtzman2019curious}. We take this critique into account by combining perplexity with a measure of language diversity based on distinct $n$-grams~\cite{li2016diversity}. Specifically, we use distinct bigrams ($\Dist{2}$) to quantify repeated two-word sequences. We define the \textit{Fluency-Diversity Rate} (FDR) ($\uparrow$) as:
\begin{equation}\label{eq:fdr}
    \text{Fluency-Diversity Rate} = \frac{(\mathrm{Dist}_2)^2}{\mathit{ln}(\mathrm{PPL})}
\end{equation}
We motivate the proposed narrativity measure with two considerations. First, we square the $\mathrm{Dist}_2$ term to impose a non-linear penalty on repetition. Because $\mathrm{Dist}_2 \in [0,1]$, squaring reduces low values disproportionately, which down-weights explanations with repeated bigram patterns even when they achieve low perplexity. This choice reduces cases where a model attains a favourable trade-off by repeating common phrases. As a result, the score is highest for text that combines lexical variety with low perplexity, a pattern associated with fluent, coherent generation~\cite{hashimoto2019unifying}. By contrast, repetitive text is penalised despite low perplexity, and incoherent text is penalised through its high perplexity in the denominator.

Second, we apply $\mathit{ln}(\cdot)$ to the perplexity term to address the mismatch between a quadratically weighted numerator and a denominator that otherwise grows exponentially with uncertainty. Since $\mathrm{PPL}=\mathit{exp}(H)$ under natural logarithms, $\mathit{ln}(\mathrm{PPL})$ is proportional to the cross-entropy $H$, so the denominator reflects average surprisal rather than its exponential transform. While a theoretically perfect model yields $\mathrm{PPL}=1$ (and thus $\mathit{ln}(\mathrm{PPL})=0$), this deterministic case is not expected in open-ended generation, which avoids division by zero in practice. To complement this, we report raw perplexity to show that predictability alone is insufficient and include perturbation analyses (shuffled, reversed, leave-one-out) to isolate structural dependencies (see Figure~\ref{fig:entropy_vs_diversity_grid}, Table~\ref{tab:summary_results} and additional analysis in Appendix~\ref{app:charts}).

\paragraph{\emph{Fluency-Diversity Rate} example.}
To compute the FDR on the examples in Table~\ref{tab:desc_vs_narr}, we combine the distinct-bigram ratios ($\text{Dist}_2 = 0.92$ for the XAI Narrative and $\text{Dist}_2 = 0.62$ for the XAI Description) with their respective perplexities ($\text{PPL} = 15.10$ and $\text{PPL} = 4.65$). Substituting into Equation~\eqref{eq:fdr}, the XAI Narrative yields $\text{FDR} = 0.92^2 / \ln(15.10) = 0.85 / 2.71 = 0.31$, whereas the XAI Description yields $\text{FDR} = 0.62^2 / \ln(4.65) = 0.38 / 1.54 = 0.25$. Although the XAI Description attains a considerably lower perplexity, the squared diversity term in the numerator penalises its repeated phrasing, so the descriptive text does not gain an advantage merely by being more predictable.

\subsection{Lexical Diversity in Narrative Explanations}
\label{subsec:features_lexical_diversity}

Our final dimension concerns the lexical diversity, which refers to the variety of vocabulary used in the explanation. Descriptive explanations often rely on repeated attribute statements, whereas narrative explanations tend to use a broader lexicon to connect events and maintain discourse coherence~\cite{weaver1991expository,graesser2011coh,saenz2002examining}.

\paragraph{Principle.}
Lexical diversity builds narrative cohesion by requiring the reader to integrate a wide range of vocabulary into a unified situation model~\cite{graesser2004coh,graesser2011coh}. Within this broad lexicon, verbs hold primary importance because they articulate the specific events and state changes that construct a continuous story, whereas descriptive lists rely heavily on static nouns and adjectives. Therefore, we emphasise the principle of lexical diversity stated by \citet{graesser2011coh}: ``\textit{Lexical diversity is related to cohesion because a higher number of different words need to be integrated into the discourse context}'' and ``\textit{(...) verbs are more prevalent in narrative discourse than in informational texts (...)}''.

\paragraph{Implementation in XAI.}
XAI Descriptions often resemble technical manuals or lists that repeat feature names and attribute values~\cite{weaver1991expository}. This repetition reduces lexical diversity and produces a static tone. In contrast, XAI Narratives use a broader vocabulary to connect factors and express relations within running text~\cite{graesser2004coh,graesser2011coh}. We emphasize the following rule: 

\begin{tcolorbox}[tightbox, title=\textbf{The lexical diversity rule}]
\setlength{\parskip}{5pt}
An XAI Narrative should use a lexically diverse vocabulary with an emphasis on active verbs to express how specific features influence the final prediction.
\end{tcolorbox}

\paragraph{Measurement.}
We quantify these properties with two metrics. First, we extend the standard \emph{Type-Token Ratio} (TTR $\uparrow$), defined as the number of unique word tokens divided by the total number of word tokens in the text, to \emph{Type-Token-adjusted Context Progression Rate} (TTCPR $\downarrow$), which is the type-token weighting of the decay rate $r$, similarly as in DCPR and other proposed metrics in this study. We define the TTCPR as:
\begin{equation}\label{eq:ttcpr}
\text{Type-Token-adjusted Context Progression Rate} = \frac{r}{\mathrm{TTR}^2}
\end{equation}
TTCPR is expected to exhibit a lower values for XAI Narratives than XAI Descriptions, as they consist of higher variety of words. We square the TTR consistent with the non-linear repetition penalty used in our DCPR and Cause-Effect metric.

\paragraph{\emph{Type-Token-adjusted Context Progression Rate} example.}
Using the decay constants $r = 0.15$ and $r = 0.13$ obtained from Equation~\eqref{eq:exp_fit} for the XAI Narrative and the XAI Description, respectively, we weigh them by the squared type-token ratios ($\text{TTR} = 0.64$ and $\text{TTR} = 0.40$). Substituting into Equation~\eqref{eq:ttcpr}, the XAI Narrative from Table~\ref{tab:desc_vs_narr} yields a TTCPR ($\downarrow$) of $0.15 / 0.64^2 = 0.37$, while the XAI Description achieves a score of $0.13 / 0.40^2 = 0.80$. The lower TTCPR for the XAI Narrative reflects its broader vocabulary, which the metric rewards through the enlarged $\text{TTR}^2$ denominator, in line with the role of lexical diversity in building narrative cohesion~\cite{graesser2011coh}.

Second, we extend the \textit{Verb Ratio} (VR $\uparrow$), defined as the proportion of verbs among all words, to \emph{Verb-adjusted Context Progression Rate} (VCPR $\downarrow$), which is verb-ratio-weighting of the decay rate $r$. We define the VCPR as:
\begin{equation}\label{eq:vcpr}
\text{Verb-adjusted Context Progression Rate} = \frac{r}{\mathrm{VR}^2}
\end{equation}
As verbs occur more frequently in narrative discourse than in informational text~\cite{graesser2011coh}, the VCPR should correctly distinguish between the XAI Narratives and XAI Descriptions. Consistent with previous approaches, we square the VR values to non-linearly penalize text with smaller verb ratios.

\paragraph{\emph{Verb-adjusted Context Progression Rate} example.}
Combining the decay constants from Equation~\eqref{eq:exp_fit} with the verb ratios ($\text{VR} = 0.104$ for the XAI Narrative and $\text{VR} = 0.086$ for the XAI Description) and substituting into Equation~\eqref{eq:vcpr}, the XAI Narrative from Table~\ref{tab:desc_vs_narr} yields a VCPR ($\downarrow$) of $0.15 / 0.104^2 = 13.87$, while the XAI Description results in a score of $0.13 / 0.086^2 = 17.58$. The lower VCPR for the XAI Narrative aligns with the observation that verbs occur more frequently in narrative discourse than in informational text~\cite{graesser2011coh}, and captures the tendency of descriptive explanations to rely on static nouns and adjectives rather than active verb constructions.

\section{Methods}
\label{sec:methods}

We compare methods that generate natural language explanations for tabular machine learning models to investigate whether they produce XAI Narratives or XAI Descriptions. Although narrative explanations exist for image and graph models~\cite{graphxain,sammani2025zeroshot}, we restrict our experiments to tabular data. To evaluate explanation as a user-centred process~\cite{bao2024understanding}, we specifically focus on multi-sentence explanations that connect feature contributions into a coherent account rather than briefly restating importance scores.

We study three state-of-the-art methods for generating natural language explanations on tabular data, specifically XAIstories~\cite{tellmeastory}, Explingo~\cite{zytek2024explingo}, and  TalkToModel~\cite{talktomodel2023explaining}. These techniques construct local explanations for individual predictions based on pre-calculated feature attributions. To establish consistent experimental conditions, we provide each method with the same set of instances to be explained and their feature attributions generated with SHAP~\cite{lundberg2017unified}. Moreover, to broaden the range of natural language generation strategies in our comparison, we introduce four extensions to existing methods: Templated Explanation, Explingo Zero-Shot, Explingo Narratives, and XAIstories Narratives. We describe the differences between these approaches in detail in the following sections. This setup supports a controlled comparison of narrative and descriptive properties using the automatic metrics.

\subsection{Datasets}
\label{subsec:datasets}

We conduct the experiments on six tabular datasets: Compas\footnote{\url{https://www.kaggle.com/datasets/danofer/compass}}, German Credit\footnote{\url{https://www.kaggle.com/datasets/uciml/german-credit/data}}, Stroke\footnote{\url{https://www.kaggle.com/datasets/fedesoriano/stroke-prediction-dataset}}, Diabetes\footnote{\url{https://www.kaggle.com/datasets/mathchi/diabetes-data-set}}, Fifa\footnote{\url{https://www.kaggle.com/datasets/mathan/fifa-2018-match-statistics}}, and Student\footnote{\url{https://www.kaggle.com/datasets/dskagglemt/student-performance-data-set}}. Spanning several domains such as risk assessment, credit scoring, clinical prediction, sports analytics, and student performance, these datasets provide the semantically meaningful features required to generate understandable natural language explanations. Additionally, their varying numbers of instances and features permit a comparison of explanation methods across different scales and input characteristics.

\begin{table*}[!htbp]
\centering
\begin{tabular}{lccc}
\toprule
Dataset & Instances & Features & Test AUC $\uparrow$ \\
\midrule
Fifa           & 128 & 24   & 0.78 \\
Student        & 395 & 33   & 0.96 \\
Diabetes       & 768 & 9    & 0.83 \\
German Credit & 1,000 & 11  & 0.67 \\
Stroke         & 5,110 & 11  & 0.85 \\
Compas         & 6,172 & 10  & 0.85 \\
\bottomrule
\end{tabular}
\caption{Dataset statistics and performance of the Random Forest classifiers. AUC is reported on the held-out test set.}
\label{tab:rf_results}
\end{table*}

To support comparable evaluation of generated natural language explanations across datasets, we require at least 30 test instances per dataset. For most datasets, we use a random 60/20/20 split into training, validation, and test sets. For the smallest dataset (Fifa), we use a fixed split of 73 training, 25 validation, and 30 test instances. From each test set, we sample 30 instances and calculate the feature attributions.

\subsection{Experimental Setup}
\label{subsec:experimental_setup}

\paragraph{Model training and validation.}
We frame each task as a supervised classification problem by training a Random Forest~\cite{breiman2001random} classifier with balanced class weights on the designated training split. Hyperparameters are tuned via grid search on the validation split across the following parameter space: number of trees ([2, 3, 4, 5, 10, 30, 50, 100, 150, 200, 250]), maximum depth ([None, 2, 3, 5, 7, 9, 11, 15]), maximum features per split ([``sqrt'', ``log2'', None]), minimum samples per split ([2, 3, 4, 6, 8, 10, 13]), and minimum samples per leaf ([1, 2, 3, 4, 6, 8, 10]). Model selection is based on the validation AUC. For most datasets, we use this validation-based selection directly. For Fifa, where the sample size is smaller, we additionally report nested cross-validation AUC to assess the stability of the selected hyperparameters. Final AUC scores on the held-out test sets appear in Table~\ref{tab:rf_results}.

\paragraph{Natural language explanation frameworks.}
Given the trained models and SHAP-based feature attributions, we generate natural language explanations using the following frameworks. For LLM-based methods, we constrain each explanation to five sentences, which typically results in text that discusses four to five of the most influential features while keeping length consistent across instances. This yields compact explanations and follows evidence that intermediate-length explanations are often perceived as most useful~\cite{domnich2025towards}. The LLM-based frameworks are prompted with a dataset description, a task description, the Random Forest prediction, the instance feature values, and the corresponding SHAP-based feature importances.

\textbf{XAIstories.}
XAIstories~\cite{tellmeastory} uses an LLM to generate multi-sentence explanations. In tabular settings, it is zero-shot prompted with dataset description, instance's feature values with corresponding feature importances, and the predicted label to produce an explanation in textual form. XAIstories is designed for narrative-style explanations and serves as a narrative-oriented baseline among the methods we compare (see Appendix~\ref{app:xaistories_prompt} for XAIstories prompt example).

\begin{xltabular}{\textwidth}{X}
\caption{Comparison of natural language explanations generated by TalkToModel~\cite{talktomodel2023explaining}, Explingo~\cite{zytek2024explingo}, and XAIstories~\cite{tellmeastory}, to explain machine learning model prediction of the instance's loan risk application from the German Credit dataset. The corresponding SHAP feature attributions for this instance are: Credit amount (+0.13), Age (-0.12), Housing (-0.04), Duration (-0.03), and Sex (-0.03).}
\label{tab:xai_examples} \\
\toprule
Generated Example of Natural Language Explanation \\
\midrule
\endfirsthead

\multicolumn{1}{c}
{{\tablename\ \thetable{} (continued from previous page)}} \\
\toprule
Generated Example of Natural Language Explanation \\
\midrule
\endhead

\multicolumn{1}{r}{{Continued on next page}} \\
\endfoot

\bottomrule
\endlastfoot

\textbf{TalkToModel~\cite{talktomodel2023explaining}}: ``For the instance 613, the model predicted good credit risk: Credit amount (Amount of credit requested) is the most important feature and has a positive influence on the prediction, Age (Age of the applicant) is the second most important feature and has a negative influence on the prediction, Housing (Housing status (Own, Rent, etc.)) is the third most important feature and has a negative influence on the prediction, Duration (Duration of the loan in months) is the 4th most important feature and has a negative influence on the prediction, Sex (Gender of the applicant) is the 5th most important feature and has a negative influence on the prediction.'' \\
\midrule
\textbf{Templated Explanation}: ``For this instance, the model predicted good credit risk. Credit amount (Amount of credit requested) is the most important feature and has a positive influence on the prediction. Age (Age of the applicant) is the second most important feature and has a negative influence on the prediction. Housing (Housing status (Own, Rent, etc.)) is the third most important feature and has a negative influence on the prediction. Duration (Duration of the loan in months) is the 4th most important feature and has a negative influence on the prediction.'' \\
\midrule
\textbf{Explingo~\cite{zytek2024explingo}}: ``The feature Credit amount (Amount of credit requested) increases the predicted value, while the features Age (Age of the applicant), Housing (Housing status (Own, Rent, etc.)), Duration (Duration of the loan in months), and Sex (Gender of the applicant) decrease the predicted value.'' \\
\midrule
\textbf{Explingo Zero-Shot}: ``The model's prediction was most strongly influenced by the amount of credit requested, which increased the likelihood of being classified as a 'bad' credit risk. The applicant's young age also contributed to a higher risk classification. Additionally, renting rather than owning a home, the loan's duration, and the applicant's gender (female) each had smaller negative impacts on the prediction, making the applicant more likely to be considered a 'bad' credit risk.'' \\ 
\midrule
\textbf{Explingo Narratives}: ``In evaluating this applicant's credit risk, the model first noted the requested credit amount of 3,632, which exerted the most significant influence by increasing the likelihood of being classified as a 'bad' credit risk. This effect was somewhat counterbalanced by the applicant's age of 22, which slightly reduced the risk, suggesting that younger applicants are viewed more favorably. However, the fact that the applicant rents their home contributed modestly to a higher risk assessment, as did the loan duration of 24 months, which nudged the prediction further toward 'bad' credit risk. Finally, the applicant's gender, being female, had a small negative impact on the risk score. Taken together, these factors interacted to shape the model's final judgment, with the requested credit amount playing the most decisive role in tipping the prediction toward a higher risk classification.'' \\ 
\midrule
\textbf{XAIstories~\cite{tellmeastory}}: ``The model predicted that this applicant is a good credit risk, and the most influential factor in this decision was the amount of credit requested, which, at 3632, is higher than the average but may signal confidence in repayment ability when considered alongside other factors. Interestingly, the applicant’s age of 22 is notably younger than most, which might typically raise concerns about limited credit history, but in this case, it did not outweigh the positive signals from other aspects. The fact that the applicant rents their home, rather than owning, could sometimes be seen as a sign of less financial stability, yet the model found this less important compared to the credit amount and age. The requested loan duration of 24 months is slightly longer than average, which could suggest a manageable repayment plan rather than a risk, especially if the applicant’s financial profile supports it. Finally, being a female applicant did not significantly sway the prediction, indicating that the model focused more on financial and situational details than on personal demographics, and overall, the combination of these factors led to a favorable assessment.'' \\
\midrule
\textbf{XAIstories Narratives}: ``For this credit applicant, the model predicted a good credit risk, and this decision was shaped most strongly by the amount of credit requested. Although the applicant asked for a higher-than-average credit amount, which might typically raise concerns, the model appears to have weighed this against other factors. Notably, the applicant’s young age could have signaled less financial experience, but in the context of the other features, it did not outweigh positive signals. The fact that the applicant rents rather than owns their home and is seeking a loan with a slightly longer duration might have introduced some caution, yet these factors were not decisive. Ultimately, the applicant’s profile, including being female, combined with the interplay of these financial and personal details, led the model to conclude that she represents a good credit risk, perhaps reflecting broader trends where responsible young renters can still be reliable borrowers.'' \\ 
\end{xltabular}

\textbf{Explingo and Explingo Zero-Shot.}
Explingo~\cite{zytek2024explingo} method follows a two-part design with a narrator that generates an explanation and a grader that evaluates it in \emph{LLM-as-a-Judge} setting. In its original form, Explingo uses few-shot prompting, where the narrator prompt includes manually-written examples of explanations. These examples are written as single long sentences, and in our experiments the model largely reproduces this format. Because single-sentence output limits narrative structure and is not suitable for sentence-level measures, we additionally introduce the \textit{Explingo Zero-Shot} variant to relax the output form (see Appendix~\ref{app:explingo_prompt} for Explingo and Explingo Zero-Shot prompt examples).

\textbf{TalkToModel and Templated Explanation.}
TalkToModel~\cite{talktomodel2023explaining} is a template-based approach in which explanations are produced by filling predefined textual templates, for instance, by listing the feature importance based on model outputs. In its original form, TalkToModel generates single-sentence explanations that present a list of influential features. This yields concise descriptive text and limits analysis of multi-sentence narrative structure.

We therefore introduce \textit{Templated Explanation}, which preserves the same template logic and placeholders as TalkToModel but presents the output as a short multi-sentence explanation. We segment the original template at feature boundaries, so that each feature is introduced in its own sentence using an ordinal framing such as ``\emph{X} is the most important feature and has a positive (negative) influence on the prediction. \emph{Y} is second most important feature and has a positive (negative) influence on the prediction.'', and so on. This retains the message of the original template while producing explanations that are more comparable in form to other methods.

\subsection{XAI Narrative Generation Rules}
\label{sub:narrative_rules}

Building on the four narrative dimensions established in Section~\ref{sec:descriptive_vs_narrative} and the conceptual definition of \emph{XAI Narratives} proposed by \citet{graphxain}, we integrate these theoretical rules directly into the prompt instructions of large language models. This approach explicitly directs the generation process to produce output with narrative properties rather than static lists of feature values. In the following, we present the exact prompt:
\begin{tcolorbox}[tightbox, title=\textbf{XAI Narrative generation rules}]
\setlength{\parskip}{5pt}
Generate a narrative explanation (an XAI Narrative) based on the following rules:
\begin{enumerate}
    \item An XAI Narrative should establish a continuous structure by following a clear narrative arc with a beginning, middle, and end, while using explicit linguistic connectives so that individual events can be seen in the perspective of the others.
    \item An XAI Narrative should explicitly identify the underlying cause-effect mechanisms to clarify why the system made a particular prediction.
    \item An XAI Narrative should explain the model's prediction with linguistic fluency, avoiding repetitive, list-like structures.
    \item An XAI Narrative should use a lexically diverse vocabulary with an emphasis on active verbs to express how specific features influence the final prediction.
\end{enumerate}
\end{tcolorbox}

These guidelines specify constraints to ensure the fluency, continuous structure, causal mechanisms, and lexical diversity of the generated explanation. We apply these generation constraints to existing baseline methods by appending the rules to their original instructions. This modification adapts the initial approaches and results in the \textit{Explingo Narratives} and \textit{XAIstories Narratives} frameworks. For all methods requiring a generative language model, the experiments are based on GPT-4.1 (\texttt{gpt-4.1-2025-04-14})~\cite{achiam2023gpt} with temperature=0.0.

\section{Results}
\label{sec:results}

\begin{figure}[!h]
  \centering
  \begin{adjustbox}{width=0.95\linewidth,center}    \includegraphics{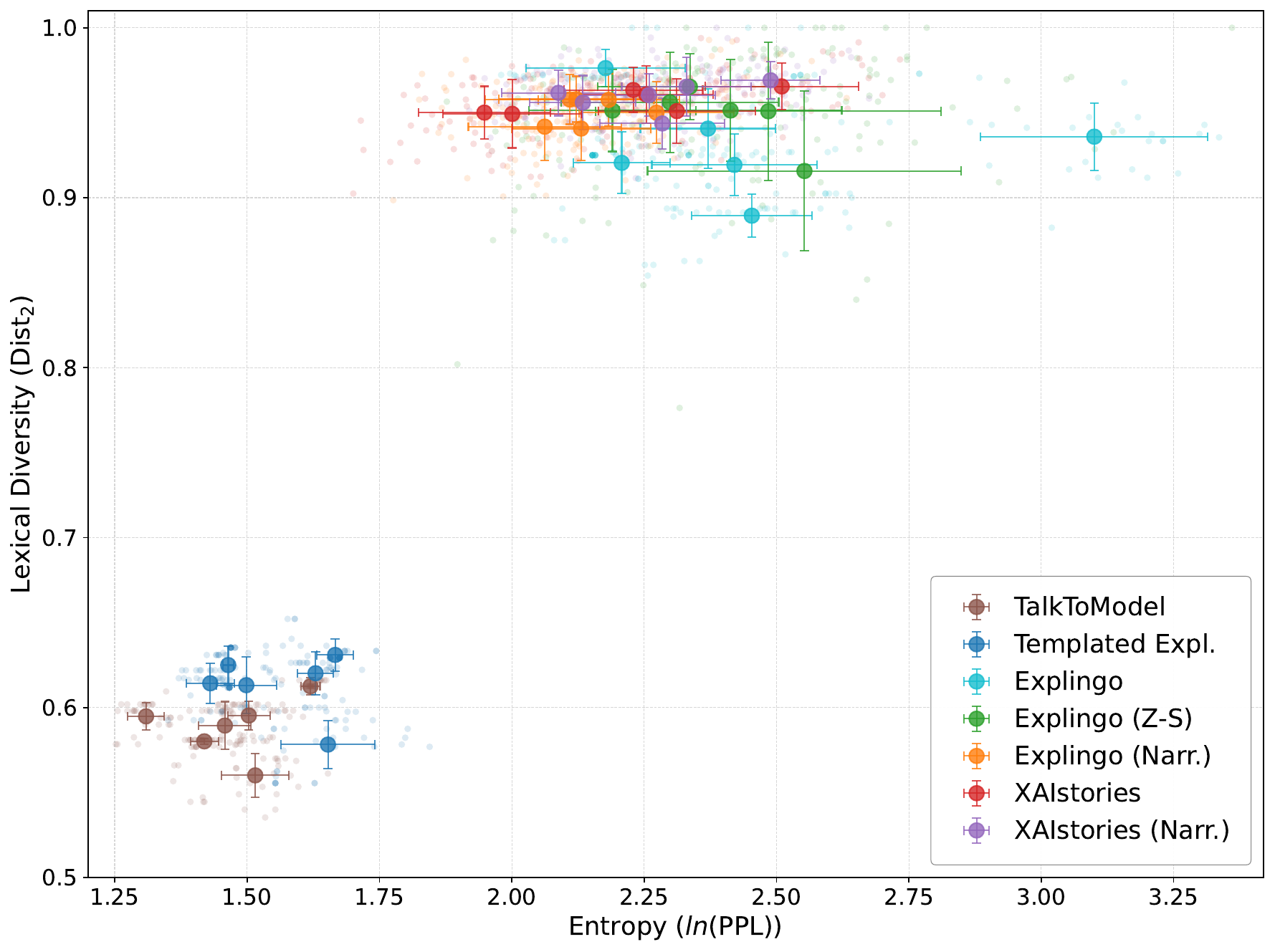}
\end{adjustbox}
    \caption{Entropy-diversity profiles of generated explanations of the templated (TalkToModel, Templated Explanations) and LLM-generated (Explingo, XAIstories) methods across datasets.}
    \label{fig:entropy_vs_diversity_grid}
\end{figure}

We begin the analysis of the differences between the explanation methods by plotting the entropy-diversity profiles, drawing on prior findings that repetitive texts tend to exhibit lower perplexity (PPL) and that narrative texts generally show greater vocabulary variability~\cite{wang2022perplexity,hashimoto2019unifying}. We use Llama 3.1-8B large language model to calculate the perplexity scores~\cite{grattafiori2024llama}. Figure~\ref{fig:entropy_vs_diversity_grid} shows explanation entropy ($\mathit{ln}(\mathrm{PPL})$) against lexical diversity across methods and datasets. Interestingly, across all datasets, template-based methods (TalkToModel and Templated Explanation) cluster in the lower-left region, combining low entropy with low diversity, which is consistent with repeated phrasing. LLM-generated methods (Explingo and XAIstories) that do not rely on fixed templates, however, fall in the upper-centre to upper-right regions, with considerably higher diversity-entropy profiles. The resulting separation shows that low perplexity can coincide with repetitive text, so perplexity alone is not an adequate quality metric to assess the text, even under the common assumption that lower PPL values indicate more predictable and thus better text~\cite{wang2022perplexity,meister2021language}.

\begin{table}[th!]
\centering
\caption{Summary of evaluation metrics for the generated explanations across six tabular datasets. Arrows in the column headers indicate the optimal direction for each metric. Bold values denote the best score for each metric across a dataset.}
\label{tab:summary_results}
\setlength{\tabcolsep}{3pt}
\renewcommand{\arraystretch}{0.85}
\resizebox{\textwidth}{!}{
    \begin{tabular}{llccccc|ccccccc}
    \toprule
    Dataset & Method & PPL $\downarrow$ & $\text{Dist}_2$ $\uparrow$ & TTR $\uparrow$ & VR $\uparrow$ & CD $\uparrow$ & \emph{FDR} $\uparrow$ & \emph{CSR} $\uparrow$ & \emph{CECPR} $\downarrow$ & \emph{DCPR} $\downarrow$ & \emph{CCPR} $\downarrow$ & \emph{TTCPR} $\downarrow$ & \emph{VCPR} $\downarrow$\\
    \midrule
    \multirow{7}{*}{Compas} & TalkToModel & \textbf{4.50} & 0.60 & 0.36 & 0.17 & 0.09 & 0.24 & - & - & - & - & - & -\\
     & Templated Expl. & 5.10 & 0.62 & 0.41 & 0.18 & 0.10 & 0.24 & 0.15 & - & 0.59 & \textbf{23.70} & 1.33 & \textbf{7.11}\\
     & Explingo & 11.70 & 0.89 & 0.62 & 0.17 & \textbf{0.14} & 0.32 & - & - & - & - & - & -\\
     & Explingo Z-S & 11.42 & 0.95 & \textbf{0.69} & \textbf{0.22} & 0.09 & 0.38 & \textbf{0.31} & - & 0.62 & 63.53 & 1.19 & 12.10\\
     & Explingo Narr. & 7.95 & 0.94 & 0.58 & 0.16 & 0.07 & \textbf{0.43} & 0.26 & 54388 & 0.50 & 82.05 & 1.17 & 14.83\\
     & XAIstories & 9.60 & \textbf{0.96} & 0.65 & 0.17 & 0.09 & 0.41 & 0.13 & 17123 & 0.44 & 55.96 & 0.97 & 13.46\\
     & XAIstories Narr. & 9.65 & \textbf{0.96} & 0.65 & 0.17 & 0.08 & 0.41 & 0.17 & \textbf{13738} & \textbf{0.42} & 55.42 & \textbf{0.91} & 12.91\\
    \midrule
    \multirow{7}{*}{Diabetes} & TalkToModel & \textbf{3.71} & 0.59 & 0.40 & 0.14 & 0.06 & 0.27 & - & - & - & - & - & -\\
     & Templated Expl. & 4.19 & 0.61 & 0.45 & 0.14 & 0.06 & 0.26 & 0.24 & - & 0.52 & 50.62 & 0.97 & 10.52\\
     & Explingo & 8.92 & \textbf{0.98} & \textbf{0.75} & 0.12 & 0.07 & 0.44 & - & - & - & - & - & -\\
     & Explingo Z-S & 10.17 & 0.96 & 0.65 & 0.15 & \textbf{0.08} & 0.40 & \textbf{0.35} & 8420 & 0.39 & 54.64 & 0.83 & 14.79\\
     & Explingo Narr. & 8.35 & 0.96 & 0.58 & 0.16 & 0.07 & 0.44 & 0.28 & 21767 & 0.40 & 66.07 & 0.87 & 11.67\\
     & XAIstories & 7.46 & 0.95 & 0.59 & \textbf{0.17} & \textbf{0.08} & \textbf{0.45} & 0.18 & \textbf{5725} & \textbf{0.28} & \textbf{39.51} & \textbf{0.72} & \textbf{8.79}\\
     & XAIstories Narr. & 8.49 & 0.96 & 0.61 & \textbf{0.17} & \textbf{0.08} & 0.43 & 0.26 & 6661 & 0.40 & 52.90 & 0.99 & 13.04\\
    \midrule
    \multirow{7}{*}{Fifa} & TalkToModel & \textbf{5.05} & 0.61 & 0.36 & 0.14 & 0.06 & 0.23 & - & - & - & - & - & -\\
     & Templated Expl. & 5.30 & 0.63 & 0.41 & 0.15 & 0.06 & 0.24 & 0.06 & - & 1.00 & 122.43 & 2.30 & 18.11\\
     & Explingo & 9.13 & 0.92 & 0.58 & 0.14 & \textbf{0.07} & 0.38 & - & - & - & - & - & -\\
     & Explingo Z-S & 9.04 & \textbf{0.95} & \textbf{0.67} & 0.14 & 0.06 & 0.41 & 0.31 & 215779 & 0.55 & 139.01 & 1.09 & 25.31\\
     & Explingo Narr. & 8.49 & 0.94 & 0.63 & 0.15 & \textbf{0.07} & \textbf{0.42} & \textbf{0.33} & 251659 & 0.67 & 126.76 & 1.49 & 26.08\\
     & XAIstories & 10.20 & \textbf{0.95} & 0.66 & \textbf{0.17} & \textbf{0.07} & 0.39 & 0.30 & \textbf{70622} & \textbf{0.42} & \textbf{67.43} & \textbf{0.86} & \textbf{13.78}\\
     & XAIstories Narr. & 9.88 & 0.94 & 0.64 & 0.16 & \textbf{0.07} & 0.39 & 0.28 & 185070 & 0.63 & 107.10 & 1.37 & 20.98\\
    \midrule
    \multirow{7}{*}{Credit} & TalkToModel & \textbf{4.30} & 0.59 & 0.35 & 0.12 & 0.06 & 0.24 & - & - & - & - & - & -\\
     & Templated Expl. & 4.48 & 0.61 & 0.40 & 0.13 & 0.06 & 0.25 & 0.18 & - & \textbf{0.38} & \textbf{42.17} & 0.89 & \textbf{8.93}\\
     & Explingo & 11.38 & 0.92 & 0.57 & 0.08 & 0.06 & 0.35 & - & - & - & - & - & -\\
     & Explingo Z-S & 13.40 & 0.92 & 0.65 & \textbf{0.17} & 0.08 & 0.33 & 0.19 & 43126 & 0.45 & 52.08 & \textbf{0.88} & 12.45\\
     & Explingo Narr. & 8.95 & 0.96 & 0.56 & 0.14 & 0.07 & \textbf{0.42} & \textbf{0.27} & 42351 & 0.44 & 54.77 & 0.96 & 15.79\\
     & XAIstories & 9.37 & 0.96 & 0.64 & 0.16 & \textbf{0.09} & \textbf{0.42} & 0.21 & 110339 & 0.41 & 47.12 & 0.94 & 15.52\\
     & XAIstories Narr. & 10.35 & \textbf{0.97} & \textbf{0.66} & 0.16 & \textbf{0.09} & 0.40 & 0.25 & \textbf{31375} & 0.44 & 47.43 & 0.94 & 15.42\\
    \midrule
    \multirow{7}{*}{Stroke} & TalkToModel & \textbf{4.14} & 0.58 & 0.36 & 0.13 & 0.07 & 0.24 & - & - & - & - & - & -\\
     & Templated Expl. & 4.33 & 0.63 & 0.43 & 0.13 & 0.07 & 0.27 & 0.24 & - & 0.58 & \textbf{43.88} & 1.19 & \textbf{13.05}\\
     & Explingo & 10.79 & 0.94 & 0.64 & 0.10 & 0.07 & 0.37 & - & - & - & - & - & -\\
     & Explingo Z-S & 10.50 & \textbf{0.97} & \textbf{0.71} & 0.14 & 0.06 & 0.40 & \textbf{0.28} & 68874 & 0.40 & 96.27 & \textbf{0.74} & 19.56\\
     & Explingo Narr. & 8.43 & 0.96 & 0.67 & \textbf{0.16} & 0.08 & 0.43 & 0.27 & 28691 & 0.45 & 68.53 & 0.92 & 16.27\\
     & XAIstories & 7.07 & 0.95 & 0.58 & 0.15 & 0.08 & \textbf{0.46} & 0.22 & 16908 & \textbf{0.35} & 54.45 & 0.93 & 13.91\\
     & XAIstories Narr. & 8.12 & 0.96 & 0.62 & 0.15 & \textbf{0.09} & 0.44 & 0.27 & \textbf{5333} & 0.48 & 59.36 & 1.14 & 19.80\\
    \midrule
    \multirow{7}{*}{Student} & TalkToModel & \textbf{4.56} & 0.56 & 0.35 & 0.15 & 0.07 & 0.21 & - & - & - & - & - & -\\
     & Templated Expl. & 5.24 & 0.58 & 0.41 & 0.15 & 0.07 & 0.20 & 0.14 & - & 0.81 & \textbf{63.36} & 1.64 & 12.84\\
     & Explingo & 22.67 & 0.94 & 0.68 & 0.12 & \textbf{0.08} & 0.28 & - & - & - & - & - & -\\
     & Explingo Z-S & 12.66 & 0.95 & \textbf{0.71} & 0.15 & \textbf{0.08} & 0.37 & 0.24 & \textbf{13665} & 0.60 & 85.17 & 1.08 & 23.71\\
     & Explingo Narr. & 9.81 & 0.95 & 0.67 & 0.14 & \textbf{0.08} & \textbf{0.40} & \textbf{0.28} & 24857 & 0.65 & 103.60 & 1.31 & 30.05\\
     & XAIstories & 12.42 & \textbf{0.97} & 0.69 & \textbf{0.17} & 0.07 & 0.37 & 0.22 & 299401 & \textbf{0.40} & 79.99 & \textbf{0.78} & \textbf{12.36}\\
     & XAIstories Narr. & 12.10 & \textbf{0.97} & 0.69 & \textbf{0.17} & \textbf{0.08} & 0.38 & 0.27 & 85370 & 0.54 & 86.44 & 1.06 & 16.71\\
    \bottomrule
    \end{tabular}
}
\end{table}

\paragraph{Standard NLP metrics can be gamed by meaningless text.}
To illustrate that the limitations of perplexity extend to the broader family of standard NLP metrics, we construct a contrastive example designed to score favourably on all of them while conveying no explanatory content:
\begin{quote}
\emph{``The model made a prediction based on the data. Therefore, the outcome is what it is, because the model made a prediction. As a result, the output follows the input, since the feature matters. Feature X is important, feature Y is less important. Consequently, the score reflects the value, so the result is the outcome. Thus, the model works, and the final decision stands.''}
\end{quote}
\noindent This text relies on tautological phrasings and high-frequency boilerplate, yet it obtains $\text{Dist}_2 = 0.94$, $\text{TTR} = 0.58$, $\text{VR} = 0.19$, and $\text{CD} = 0.13$, most of which match or exceed the scores of the XAI Narrative in Table~\ref{tab:desc_vs_narr}. Its perplexity ($17.37$) is only moderately higher than that of the XAI Narrative ($15.10$), despite the text being obviously meaningless as an explanation. This example shows that lexical diversity ($\mathrm{Dist}_2$), type-token ratio (TTR), verb ratio (VR), and connective density (CD) can all be inflated independently of whether the text actually explains a prediction, which further motivates the need for metrics that capture narrative structure.

In Table~\ref{tab:summary_results}, we report our main results. Across standard NLP metrics, including perplexity, distinct bigram ratio, type-token ratio, verb ratio, and connectives density, the patterns in each method vary across datasets and do not show a clear and consistent separation between methods. The results also reiterate that perplexity alone is not a sufficient quality metric. TalkToModel and Templated Explanation, which attain the lowest perplexity, produce templated, repetitive descriptions that provide limited justification for why specific features support the prediction.

Lexical and discourse markers alone show clear limitations as quality metrics. The distinct bigram ratio, type-token ratio, verb ratio, and connective density usually favour template-free outputs, but they cannot serve as standalone measures, since maximising them produces incoherent text composed of non-repetitive words without overall meaning. The proposed \emph{Narrativity} metrics (FDR, CSR, CECPR, DCPR, CCPR, TTCPR, and VCPR) give a more informative assessment and, across datasets, assign worse scores to template-based methods than to template-free ones, separating XAI Narratives from XAI Descriptions more reliably than standard NLP metrics. Since all of the Context Progression Rate metrics depend on the decay constant $r$ from the shifted exponential in Equation (2), their reliability rests on the quality of the fit. The goodness-of-fit statistics reported in Table~\ref{tab:fitted_dcpr} (see Appendix~\ref{app:fitting_table}) show $R^2$ values at or above 0.98 across all datasets and methods, with RMSE below 1 perplexity unit in most settings. The largest residuals appear for Templated Explanation on Student dataset (RMSE = 1.71), Explingo Narratives on Diabetes dataset (RMSE = 1.51), and Explingo Zero-Shot on Stroke dataset (RMSE = 1.41), yet all three retain $R^2 \ge 0.98$, indicating that $r$ reflects genuine context progression rather than fitting artefacts. Moreover, Figure~\ref{fig:cumulaive_perplexity_frameworks} (Appendix~\ref{app:charts}) shows the cumulative sentence perplexity trajectories which clearly follow the shifted exponential curve and further reinforce the relevance of our approach.

\begin{figure}[h!] 
    \centering 
    \begin{adjustbox}{width=1\linewidth,center} 
        \includegraphics{plots/rank_heatmap_all.pdf} 
    \end{adjustbox} 
    \caption{Average framework ranks across metrics and six datasets. Each cell shows the mean rank for a framework-metric pair (1 = best, 7 = worst). Rows are sorted by overall performance. Color scale: green (best), red (worst). Missing values assigned equal last rank per block.} 
    \label{fig:avg_ranks} 
\end{figure}

\begin{figure}[t!]
    \centering
    \begin{adjustbox}{width=1\linewidth,center} 
        \includegraphics{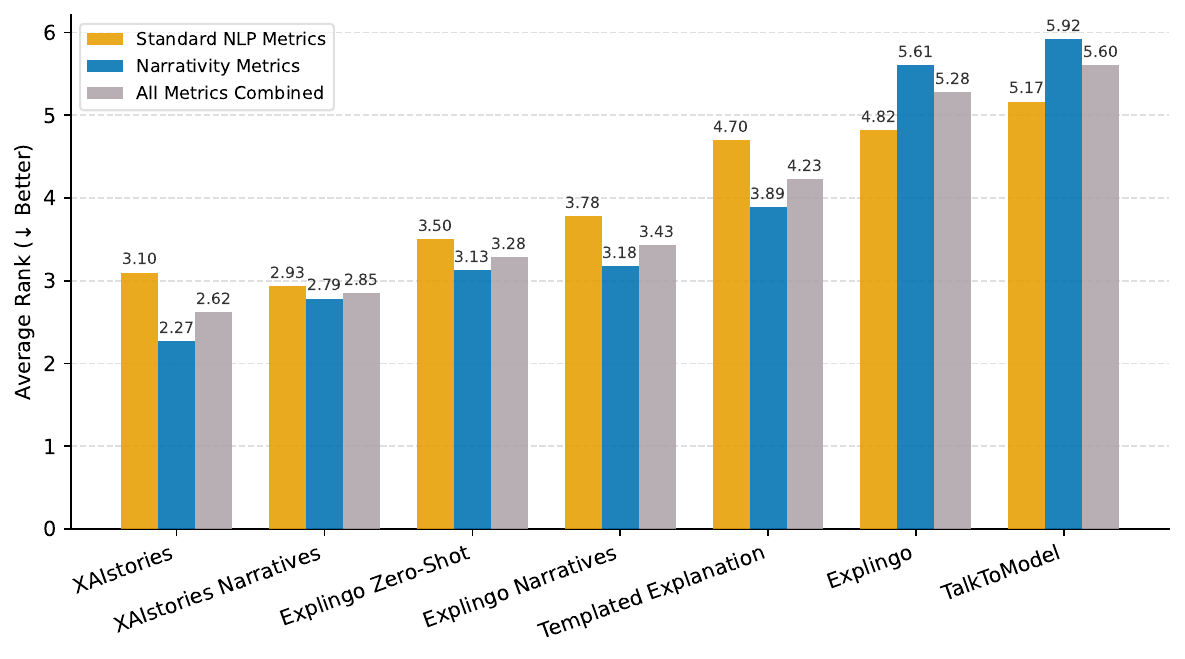} 
    \end{adjustbox} 
    \caption{Framework average ranks across metric groups. Bars represent mean of per-metric average ranks (lower = better). Standard NLP Metrics (PPL, Dist\textsubscript{2}, TTR, VR, CD) are colored in orange. \emph{Narrativity} Metrics (FDR, CSR, CECPR, DCPR, CCPR, TTCPR, VCPR) are colored in blue. All Metrics Combined (gray) account for all metrics. Frameworks are ordered by overall performance.} 
    \label{fig:rank_by_group} 
\end{figure}

Figure~\ref{fig:avg_ranks} presents the result of average ranks aggregated over six datasets. Methods where results are missing (due to absence of e.g., cause-effect markers or text separation into sentences) are given equal lowest ranks. Whereas for standard NLP metrics we see confusing results (e.g., TalkToModel and Templated Explanation achieve the best ranks for PPL and simultaneously score lowest ranks for other metrics), the consistency of average ranks over the proposed seven metrics is clear. Although overall results of averaged ranks over 12 metrics show that XAIstories and XAIstories Narratives methods result in the highest averaged ranks, outperforming Explingo and TalkToModel methods that score the worst, even more clear separation is observed when only seven proposed metrics are investigated. The average rank results with distinction between the standard NLP metrics and the proposed \emph{narrativity} metrics are shown in Figure~\ref{fig:rank_by_group}. Interestingly, there is not much difference in the average ranks between XAIstories and the version extended by the XAI Narrative generation rules. This is perhaps due to the fact that the original XAIstories method already indicates in its prompt to generate a narrative to explain model's prediction. Nevertheless, we observe substantial differences between the average rank results between the Explingo and both the Zero-Shot and Narratives versions. The original Explingo method includes examples of explanations in its prompt (few shot approach), resulting in the new generated explanations to strictly follow the provided examples, leaving little room for the LLM to generate a narrative explanation.

\subsection{Statistical comparison}

We continue the analysis with the non-parametric statistical test comparison. We follow the framework proposed by~\citet{demvsar2006statistical} for comparing the results over multiple datasets. Prior to pairwise comparisons, we apply the Friedman omnibus test for each metric at $\alpha=0.05$. This test determines whether any statistical differences exist in the ranks among the frameworks. The null hypothesis assumes that all frameworks share the same expected rank. Table~\ref{tab:friedman_summary} presents these results. Eleven of the twelve metrics show $p<0.05$. The connectives density (CD) metric yields $p=0.10$, so we conduct no post-hoc analysis for that specific measure. For the eleven metrics showing a significant Friedman test result, we perform the Nemenyi post-hoc test to determine which specific pairs of frameworks differ from one another. The formula defines the required difference threshold, critical difference, as:
\begin{equation}
\text{Critical Difference} = q_\alpha\sqrt{\frac{k(k+1)}{6N}},
\end{equation}
where $q_\alpha$ is the distribution value from the studentized range distribution divided by the square root of two, $k$ denotes the number of compared methods (frameworks), and $N$ refers to the number of blocks (datasets). With $k=7$ and $N=6$, the resulting critical difference is approximately 3.68. Since the number of frameworks is large relative to the number of datasets, this threshold requires a difference of nearly 3.7 rank units on a 7-point scale to indicate the statistical significance between methods. This makes the Nemenyi test conservative by design. Table~\ref{tab:nemenyi_posthoc} reports the pairwise $p$-values for all 21 framework combinations across eleven metrics. Cells are shaded in green whenever a pairwise contrast is statistically significant, with darker shades indicating stronger evidence: the lightest shade marks $p<0.05$, the medium shade marks $p<0.01$, and the darkest shade marks $p<0.001$.

\begin{table}[!t]
\centering
\caption{Friedman test summary. For each metric, tests the null 
hypothesis that all frameworks have equal average ranks. Methods with 
missing results receive an equal last-rank penalty. \newline Note: $^{***}\,p<0.001$, $^{**}\,p<0.01$, $^{*}\,p<0.05$.}
\label{tab:friedman_summary}
\setlength{\tabcolsep}{5pt}
\renewcommand{\arraystretch}{1.0}
\newcommand{\sig}[1]{\makebox[1.8em][l]{\footnotesize$^{#1}$}}
\begin{tabular}{lccr@{}l}
    \toprule
    Metric & $\chi^2$ & $df$ & \multicolumn{2}{c}{$p$-value} \\
    \midrule
    PPL      & 30.21 & 6 & $<0.001$ & \sig{***} \\
    Dist$_2$ & 27.28 & 6 & $<0.001$ & \sig{***} \\
    TTR      & 29.68 & 6 & $<0.001$ & \sig{***} \\
    VR       & 18.73 & 6 & $0.005$  & \sig{**}  \\
    CD       & 10.69 & 6 & $0.099$  & \sig{}    \\
    FDR      & 29.93 & 6 & $<0.001$ & \sig{***} \\
    CSR      & 31.75 & 6 & $<0.001$ & \sig{***} \\
    CECPR    & 29.37 & 6 & $<0.001$ & \sig{***} \\
    DCPR     & 28.90 & 6 & $<0.001$ & \sig{***} \\
    CCPR     & 33.09 & 6 & $<0.001$ & \sig{***} \\
    TTCPR    & 28.10 & 6 & $<0.001$ & \sig{***} \\
    VCPR     & 29.75 & 6 & $<0.001$ & \sig{***} \\
    \bottomrule
\end{tabular}
\end{table}

\begin{table}[!htb]
\centering
\caption{Nemenyi post-hoc pairwise $p$-values. Cells with $p < 0.05$ are highlighted in green. Methods with missing results on a dataset receive an equal last rank penalty. Note: $^{***}\,p<0.001$, $^{**}\,p<0.01$, $^{*}\,p<0.05$.}
\label{tab:nemenyi_posthoc}
\setlength{\tabcolsep}{3pt}
\renewcommand{\arraystretch}{0.9}
\resizebox{\textwidth}{!}{
    \begin{tabular}{lccccc|ccccccc}
    \toprule
    Framework pair & PPL & Dist$_2$ & TTR & VR & CD & FDR & CSR & CECPR & DCPR & CCPR & TTCPR & VCPR \\
    \midrule
    TalkToModel vs Explingo & \cellcolor[HTML]{4CAF50}$<$0.001$^{***}$ & 0.293 & 0.074 & 0.967 & -- & 0.591 & 1.000 & 1.000 & 1.000 & 1.000 & 1.000 & 1.000 \\
    TalkToModel vs Templ. Expl. & 0.985 & 0.985 & 0.985 & 0.985 & -- & 1.000 & 0.893 & 1.000 & 0.763 & \cellcolor[HTML]{90EE90}0.006$^{**}$ & 0.835 & \cellcolor[HTML]{90EE90}0.006$^{**}$ \\
    TalkToModel vs XAIstories & 0.105 & \cellcolor[HTML]{90EE90}0.009$^{**}$ & \cellcolor[HTML]{DFF5DC}0.035$^{*}$ & 0.196 & -- & \cellcolor[HTML]{90EE90}0.007$^{**}$ & 0.763 & 0.412 & \cellcolor[HTML]{90EE90}0.003$^{**}$ & \cellcolor[HTML]{DFF5DC}0.015$^{*}$ & \cellcolor[HTML]{DFF5DC}0.018$^{*}$ & \cellcolor[HTML]{DFF5DC}0.035$^{*}$ \\
    TalkToModel vs XAIstories Narr. & \cellcolor[HTML]{DFF5DC}0.035$^{*}$ & \cellcolor[HTML]{90EE90}0.002$^{**}$ & \cellcolor[HTML]{90EE90}0.009$^{**}$ & 0.258 & -- & \cellcolor[HTML]{DFF5DC}0.042$^{*}$ & 0.124 & 0.145 & 0.145 & 0.145 & 0.226 & 0.412 \\
    TalkToModel vs Explingo Z-S & \cellcolor[HTML]{90EE90}0.001$^{**}$ & \cellcolor[HTML]{DFF5DC}0.012$^{*}$ & \cellcolor[HTML]{4CAF50}$<$0.001$^{***}$ & 0.723 & -- & 0.330 & \cellcolor[HTML]{DFF5DC}0.023$^{*}$ & 0.763 & 0.258 & 0.763 & \cellcolor[HTML]{DFF5DC}0.035$^{*}$ & 0.412 \\
    TalkToModel vs Explingo Narr. & 0.412 & \cellcolor[HTML]{DFF5DC}0.035$^{*}$ & 0.330 & 0.967 & -- & \cellcolor[HTML]{90EE90}0.002$^{**}$ & \cellcolor[HTML]{90EE90}0.007$^{**}$ & 0.763 & 0.412 & 0.937 & 0.412 & 0.835 \\
    Explingo vs Templ. Expl. & \cellcolor[HTML]{90EE90}0.009$^{**}$ & 0.800 & 0.412 & 0.591 & -- & 0.680 & 0.893 & 1.000 & 0.763 & \cellcolor[HTML]{90EE90}0.006$^{**}$ & 0.835 & \cellcolor[HTML]{90EE90}0.006$^{**}$ \\
    Explingo vs XAIstories & 0.680 & 0.866 & 1.000 & \cellcolor[HTML]{DFF5DC}0.015$^{*}$ & -- & 0.545 & 0.763 & 0.412 & \cellcolor[HTML]{90EE90}0.003$^{**}$ & \cellcolor[HTML]{DFF5DC}0.015$^{*}$ & \cellcolor[HTML]{DFF5DC}0.018$^{*}$ & \cellcolor[HTML]{DFF5DC}0.035$^{*}$ \\
    Explingo vs XAIstories Narr. & 0.893 & 0.636 & 0.994 & \cellcolor[HTML]{DFF5DC}0.023$^{*}$ & -- & 0.866 & 0.124 & 0.145 & 0.145 & 0.145 & 0.226 & 0.412 \\
    Explingo vs Explingo Z-S & 1.000 & 0.893 & 0.591 & 0.169 & -- & 1.000 & \cellcolor[HTML]{DFF5DC}0.023$^{*}$ & 0.763 & 0.258 & 0.763 & \cellcolor[HTML]{DFF5DC}0.035$^{*}$ & 0.412 \\
    Explingo vs Explingo Narr. & 0.258 & 0.977 & 0.994 & 0.500 & -- & 0.293 & \cellcolor[HTML]{90EE90}0.007$^{**}$ & 0.763 & 0.412 & 0.937 & 0.412 & 0.835 \\
    Templ. Expl. vs XAIstories & 0.500 & 0.105 & 0.258 & 0.680 & -- & \cellcolor[HTML]{DFF5DC}0.012$^{*}$ & 1.000 & 0.412 & 0.258 & 1.000 & 0.455 & 0.998 \\
    Templ. Expl. vs XAIstories Narr. & 0.258 & \cellcolor[HTML]{DFF5DC}0.035$^{*}$ & 0.105 & 0.763 & -- & 0.062 & 0.800 & 0.145 & 0.937 & 0.937 & 0.954 & 0.680 \\
    Templ. Expl. vs Explingo Z-S & \cellcolor[HTML]{DFF5DC}0.023$^{*}$ & 0.124 & \cellcolor[HTML]{90EE90}0.003$^{**}$ & 0.990 & -- & 0.412 & 0.412 & 0.763 & 0.985 & 0.330 & 0.591 & 0.680 \\
    Templ. Expl. vs Explingo Narr. & 0.893 & 0.258 & 0.835 & 1.000 & -- & \cellcolor[HTML]{90EE90}0.003$^{**}$ & 0.226 & 0.763 & 0.998 & 0.145 & 0.994 & 0.258 \\
    XAIstories vs XAIstories Narr. & 1.000 & 1.000 & 1.000 & 1.000 & -- & 0.998 & 0.917 & 0.998 & 0.893 & 0.985 & 0.967 & 0.937 \\
    XAIstories vs Explingo Z-S & 0.835 & 1.000 & 0.763 & 0.977 & -- & 0.800 & 0.591 & 0.998 & 0.763 & 0.500 & 1.000 & 0.937 \\
    XAIstories vs Explingo Narr. & 0.994 & 1.000 & 0.967 & 0.763 & -- & 1.000 & 0.370 & 0.998 & 0.591 & 0.258 & 0.866 & 0.591 \\
    XAIstories Narr. vs Explingo Z-S & 0.967 & 0.999 & 0.937 & 0.990 & -- & 0.977 & 0.997 & 0.937 & 1.000 & 0.937 & 0.990 & 1.000 \\
    XAIstories Narr. vs Explingo Narr. & 0.937 & 0.985 & 0.835 & 0.835 & -- & 0.967 & 0.967 & 0.937 & 0.998 & 0.763 & 1.000 & 0.994 \\
    Explingo Z-S vs Explingo Narr. & 0.412 & 1.000 & 0.196 & 0.997 & -- & 0.545 & 1.000 & 1.000 & 1.000 & 1.000 & 0.937 & 0.994 \\
    \bottomrule
    \end{tabular}
}
\end{table}

The Nemenyi post-hoc comparisons show a consistent separation between templated and LLM-based methods, although the conservative critical difference of roughly 3.68 rank units on a 7-point scale limits the number of pairs reaching significance at $\alpha$ = 0.05 given only six datasets. The clearest contrasts arise between TalkToModel and the LLM-generated approaches, while pairwise differences among the three strongest methods, namely XAIstories, XAIstories Narratives, and Explingo Narratives, remain non-significant on every metric. This pattern suggests that once a method produces multi-sentence explanations with narrative intent, the outputs converge on comparable narrativity scores across the proposed metrics. The CECPR metric yields no significant pairwise differences, with the closest to significance being TalkToModel and Templated Explanation against XAIstories Narratives, both at $p=0.15$.

\paragraph{NLP vs. \emph{Narrativity} metrics comparison.} The results presented in Table~\ref{tab:nemenyi_posthoc} show that across the standard NLP metrics, significant differences are scattered and do not form a coherent ranking. TalkToModel separates from XAIstories Narratives on PPL, Dist$_2$, and TTR, and from Explingo Zero-Shot on PPL and TTR, but these separations partly reflect the low baseline perplexity of templated text rather than higher quality of the generated explanation. The proposed \emph{narrativity} metrics instead produce a more systematic picture. TalkToModel differs from XAIstories on FDR, DCPR, CCPR, TTCPR, and VCPR, and Explingo differs from XAIstories on DCPR, CCPR, TTCPR, and VCPR, which indicates that the \emph{narrativity} metrics recover the templated versus narrative distinction more reliably than standard NLP metrics.

\paragraph{Effect of the narrative generation rules.} The impact of the XAI Narrative generation rules appears most clearly for Explingo. Explingo Narratives differs significantly from the original Explingo on CSR, showing that appending the rules changes the order-dependent structure of the generated text. In contrast, XAIstories and XAIstories Narratives do not differ significantly on any metric, which aligns with the observation that the original XAIstories prompt already instructs the model to generate a narrative. The rules therefore introduce narrative structure where the baseline prompt does not encourage it, and leave outputs largely unchanged where narrative intent is already present.

Beyond the main results reported above, Appendix~\ref{app:charts} presents additional perturbation analyses that further support the distinction between templated and narrative explanations. Figure~\ref{fig:ppl_relative_increase} reports the nominal change in perplexity under shuffled, reversed, and leave-one-out reordering across the six datasets, while Figure~\ref{fig:cumulaive_perplexity_sentence_order} compares the cumulative perplexity trajectories of Templated Explanations and XAIstories Narratives under shuffling and reversal. In both cases, only the narrative outputs exhibit order-dependent behaviour. Finally, Figure~\ref{fig:cumulaive_perplexity_distribution} shows that single-sentence ablation reveals stronger inter-sentence dependence in narrative explanations than in templated text.

\section{Conclusion}
\label{sec:conclusion}

In this work, we examine methods that generate natural language explanations and demonstrate that most of them result primarily in descriptive explanations. Although these approaches successfully identify the features that influence a prediction, they provide little insight into \emph{why} the prediction occurs. Drawing on findings from social sciences and linguistics, we emphasise the need for explanations in the form of \emph{XAI Narratives} as a way to explain the relations between features in a manner that aligns with how readers integrate information across sentences. Drawing on our findings, we provide the set of four rules that every natural language explanation should have to result in an XAI Narrative that explains not only which features are important for the final prediction, but also \emph{why} they are important.

Our empirical analysis shows that standard NLP measures such as perplexity, lexical diversity, or keyword counts do not reliably distinguish descriptive, list-like explanations from narrative ones. Template-based methods can attain low perplexity through repeated phrasing, and a meaningless tautological text can match the lexical diversity, type-token ratio, verb density, and connective density of a genuine narrative, while offering no explanatory structure. To address this issue, we introduce seven automatic metrics that quantify the narrativity of an explanation along four dimensions: fluency (Fluency-Diversity Rate), continuous structure (Continuous Structure Rate, Diversity-adjusted Context Progression Rate, Connectives-adjusted Context Progression Rate), cause-effect mechanisms (Cause-Effect-adjusted Context Progression Rate), and lexical diversity (Type-Token-adjusted Context Progression Rate, Verb-adjusted Context Progression Rate).

We benchmark TalkToModel, Explingo, and XAIstories, together with several extensions we introduce to these methods (Templated Explanation, Explingo Zero-Shot, Explingo Narratives, and XAIstories Narratives), on explanations generated for model predictions across six tabular datasets. The proposed metrics separate narrative from descriptive explanation styles more reliably than standard NLP measures, and non-parametric tests confirm that the resulting ranking of methods is statistically consistent across datasets. We further show that the proposed Narrative Generation Rules improve the narrativity of the produced explanations, an effect most clearly observed in the Explingo Narratives framework. For XAIstories, the same rules yield only negligible changes, since its original prompt already instructs the model to generate narrative explanations.

We discuss several limitations that suggest directions for further work. First, the empirical evaluation focuses on tabular data. Tabular models are common in high-stakes settings where the form of the explanation matters, but this choice excludes other modalities. Second, components based on perplexity can be sensitive to tokenisation and punctuation, which can introduce variability between base language models and between texts~\cite{wang2022perplexity}. Third, although the proposed metrics are motivated by findings from social sciences and linguistics, the evaluation relies on automated measures rather than large-scale human assessment.

Future work can extend the evaluation to non-tabular domains, including images and graphs, to test whether the metrics transfer to settings with different input structures and explanation conventions. Controlled user studies can also examine how metrics relate to comprehension, mental model formation, decision accuracy, and whether higher scores align with improved understanding. In addition, comparisons with broader narrative corpora such as StoryDB~\cite{tikhonov2021storydb} may help separate the properties of general storytelling from explanation-specific narratives. Finally, hybrid measures that combine entropy-based signals with discourse and content markers may provide a more detailed view of the explanatory structure.

\section*{Data and Code Availability}
The data and code used in this research, are available at: \url{https://github.com/ADMAntwerp/On-the-Importance-and-Evaluation-of-Narrativity-in-Natural-Language-AI-Explanations}

\section*{Acknowledgments}
This research received funding from the Flemish Government under the “Onderzoeksprogramma Artifici\"{e}le Intelligentie (AI) Vlaanderen”.

\bibliographystyle{unsrtnat}
\bibliography{bib}

\begin{thebibliography}{71}
\providecommand{\natexlab}[1]{#1}
\providecommand{\url}[1]{\texttt{#1}}
\expandafter\ifx\csname urlstyle\endcsname\relax
  \providecommand{\doi}[1]{doi: #1}\else
  \providecommand{\doi}{doi: \begingroup \urlstyle{rm}\Url}\fi

\bibitem[Miller(2019)]{miller2019explanation}
Tim Miller.
\newblock Explanation in artificial intelligence: Insights from the social sciences.
\newblock \emph{Artificial intelligence}, 267:\penalty0 1--38, 2019.

\bibitem[Martens et~al.(2025)Martens, Hinns, Dams, Vergouwen, and Evgeniou]{tellmeastory}
David Martens, James Hinns, Camille Dams, Mark Vergouwen, and Theodoros Evgeniou.
\newblock Tell me a story! {N}arrative-driven {XAI} with {L}arge {L}anguage {M}odels.
\newblock \emph{Decision Support Systems}, 191:\penalty0 114402, 2025.
\newblock ISSN 0167-9236.
\newblock \doi{https://doi.org/10.1016/j.dss.2025.114402}.
\newblock URL \url{https://www.sciencedirect.com/science/article/pii/S016792362500003X}.

\bibitem[Miller et~al.(2017)Miller, Howe, and Sonenberg]{miller2017explainable}
Tim Miller, Piers Howe, and Liz Sonenberg.
\newblock Explainable ai: Beware of inmates running the asylum or: How i learnt to stop worrying and love the social and behavioural sciences.
\newblock \emph{arXiv preprint arXiv:1712.00547}, 2017.

\bibitem[Dahlstrom(2014)]{dahlstrom2014using}
Michael~F Dahlstrom.
\newblock Using narratives and storytelling to communicate science with nonexpert audiences.
\newblock \emph{Proceedings of the national academy of sciences}, 111\penalty0 (supplement\_4):\penalty0 13614--13620, 2014.

\bibitem[Cedro and Martens(2025)]{graphxain}
Mateusz Cedro and David Martens.
\newblock Graph{XAIN}: Narratives to {E}xplain {G}raph {N}eural {N}etworks.
\newblock In \emph{Explainable Artificial Intelligence}, pages 91--114. Springer Nature Switzerland, 2025.
\newblock ISBN 978-3-032-08327-2.

\bibitem[Zytek et~al.(2024)Zytek, Pido, Alnegheimish, Berti-Equille, and Veeramachaneni]{zytek2024explingo}
Alexandra Zytek, Sara Pido, Sarah Alnegheimish, Laure Berti-Equille, and Kalyan Veeramachaneni.
\newblock Explingo: Explaining ai predictions using large language models.
\newblock In \emph{2024 IEEE International Conference on Big Data (BigData)}, pages 1197--1208. IEEE, 2024.

\bibitem[Slack et~al.(2023)Slack, Krishna, Lakkaraju, and Singh]{talktomodel2023explaining}
Dylan Slack, Satyapriya Krishna, Himabindu Lakkaraju, and Sameer Singh.
\newblock Explaining machine learning models with interactive natural language conversations using talktomodel.
\newblock \emph{Nature Machine Intelligence}, 5\penalty0 (8):\penalty0 873--883, 2023.

\bibitem[Friedman(1974)]{friedman1974explanation}
Michael Friedman.
\newblock Explanation and scientific understanding.
\newblock \emph{the Journal of Philosophy}, 71\penalty0 (1):\penalty0 5--19, 1974.

\bibitem[Norris et~al.(2005)Norris, Guilbert, Smith, Hakimelahi, and Phillips]{norris2005theoretical}
Stephen~P Norris, Sandra~M Guilbert, Martha~L Smith, Shahram Hakimelahi, and Linda~M Phillips.
\newblock A theoretical framework for narrative explanation in science.
\newblock \emph{Science education}, 89\penalty0 (4):\penalty0 535--563, 2005.

\bibitem[Bruner(1986)]{bruner2009actual}
Jerome~S Bruner.
\newblock \emph{Actual minds, possible worlds}.
\newblock Harvard university press, 1986.

\bibitem[Lundberg and Lee(2017)]{lundberg2017unified}
Scott~M Lundberg and Su-In Lee.
\newblock A unified approach to interpreting model predictions.
\newblock \emph{Advances in neural information processing systems}, 30, 2017.

\bibitem[Ribeiro et~al.(2016)Ribeiro, Singh, and Guestrin]{ribeiro2016should}
Marco~Tulio Ribeiro, Sameer Singh, and Carlos Guestrin.
\newblock "{W}hy should {I} trust you?" {E}xplaining the predictions of any classifier.
\newblock In \emph{Proceedings of the 22nd ACM SIGKDD international conference on knowledge discovery and data mining}, pages 1135--1144, 2016.

\bibitem[Martens and Provost(2014)]{martens2014explaining}
David Martens and Foster Provost.
\newblock Explaining data-driven document classifications.
\newblock \emph{MIS quarterly}, 38\penalty0 (1):\penalty0 73--100, 2014.

\bibitem[Simonyan et~al.(2013)Simonyan, Vedaldi, and Zisserman]{simonyan2013deep}
Karen Simonyan, Andrea Vedaldi, and Andrew Zisserman.
\newblock Deep inside convolutional networks: Visualising image classification models and saliency maps.
\newblock \emph{arXiv preprint arXiv:1312.6034}, 2013.

\bibitem[Selvaraju et~al.(2017)Selvaraju, Cogswell, Das, Vedantam, Parikh, and Batra]{Selvaraju2017Grad_cam}
Ramprasaath~R. Selvaraju, Michael Cogswell, Abhishek Das, Ramakrishna Vedantam, Devi Parikh, and Dhruv Batra.
\newblock Grad-cam: Visual explanations from deep networks via gradient-based localization.
\newblock In \emph{2017 IEEE International Conference on Computer Vision (ICCV)}, pages 618--626, 2017.
\newblock \doi{10.1109/ICCV.2017.74}.

\bibitem[Ying et~al.(2019)Ying, Bourgeois, You, Zitnik, and Leskovec]{ying2019gnnexplainer}
Zhitao Ying, Dylan Bourgeois, Jiaxuan You, Marinka Zitnik, and Jure Leskovec.
\newblock Gnnexplainer: Generating explanations for graph neural networks.
\newblock \emph{Advances in neural information processing systems}, 32, 2019.

\bibitem[Cambria et~al.(2023)Cambria, Malandri, Mercorio, Mezzanzanica, and Nobani]{cambria2023survey}
Erik Cambria, Lorenzo Malandri, Fabio Mercorio, Mario Mezzanzanica, and Navid Nobani.
\newblock A survey on xai and natural language explanations.
\newblock \emph{Information Processing \& Management}, 60\penalty0 (1):\penalty0 103111, 2023.

\bibitem[Li et~al.(2018)Li, Pan, Wang, Yang, and Cambria]{LI2018301}
Yang Li, Quan Pan, Suhang Wang, Tao Yang, and Erik Cambria.
\newblock A generative model for category text generation.
\newblock \emph{Information Sciences}, 450:\penalty0 301--315, 2018.
\newblock ISSN 0020-0255.
\newblock \doi{https://doi.org/10.1016/j.ins.2018.03.050}.
\newblock URL \url{https://www.sciencedirect.com/science/article/pii/S0020025518302366}.

\bibitem[Pan et~al.(2025)Pan, Xiong, Wu, Zhang, Zhang, Hu, and Zhao]{pan2025graphnarrator}
Bo~Pan, Zhen Xiong, Guanchen Wu, Zheng Zhang, Yifei Zhang, Yuntong Hu, and Liang Zhao.
\newblock Graphnarrator: Generating textual explanations for graph neural networks.
\newblock In \emph{Proceedings of the 63rd Annual Meeting of the Association for Computational Linguistics (Volume 1: Long Papers)}, pages 23--42, 2025.

\bibitem[Sammani and Deligiannis(2025)]{sammani2025zeroshot}
Fawaz Sammani and Nikos Deligiannis.
\newblock Zero-shot natural language explanations.
\newblock In \emph{The Thirteenth International Conference on Learning Representations}, 2025.
\newblock URL \url{https://openreview.net/forum?id=X6VVK8pIzZ}.

\bibitem[He and Martens(2026)]{he2026agentic}
Yifan He and David Martens.
\newblock An agentic approach to generating {XAI}-{N}arratives.
\newblock \emph{arXiv preprint arXiv:2603.20003}, 2026.

\bibitem[Zhou et~al.(2023)Zhou, Zhang, and Tan]{zhou2023}
Yangqiaoyu Zhou, Yiming Zhang, and Chenhao Tan.
\newblock {FL}am{E}: Few-shot learning from natural language explanations.
\newblock In \emph{Proceedings of the 61st Annual Meeting of the Association for Computational Linguistics (Volume 1: Long Papers)}, pages 6743--6763. Association for Computational Linguistics, 2023.
\newblock \doi{10.18653/v1/2023.acl-long.372}.
\newblock URL \url{https://aclanthology.org/2023.acl-long.372/}.

\bibitem[Mariotti et~al.(2020)Mariotti, Alonso, and Gatt]{mariotti2020towards}
Ettore Mariotti, Jose~M Alonso, and Albert Gatt.
\newblock Towards harnessing natural language generation to explain black-box models.
\newblock In \emph{2nd Workshop on interactive natural language technology for explainable artificial intelligence}, pages 22--27, 2020.

\bibitem[Jelinek et~al.(1977)Jelinek, Mercer, Bahl, and Baker]{jelinek1977perplexity}
Fred Jelinek, Robert~L Mercer, Lalit~R Bahl, and James~K Baker.
\newblock Perplexity—a measure of the difficulty of speech recognition tasks.
\newblock \emph{The Journal of the Acoustical Society of America}, 62\penalty0 (S1):\penalty0 S63--S63, 1977.

\bibitem[Ichmoukhamedov et~al.(2024)Ichmoukhamedov, Hinns, and Martens]{ichmoukhamedov2024good}
Timour Ichmoukhamedov, James Hinns, and David Martens.
\newblock How good is my story? {T}owards quantitative metrics for evaluating {LLM}-generated {XAI} narratives.
\newblock \emph{arXiv preprint arXiv:2412.10220}, 2024.

\bibitem[Shirvani-Mahdavi and Li(2025)]{shirvani2025rule2text}
Nasim Shirvani-Mahdavi and Chengkai Li.
\newblock Rule2text: A framework for generating and evaluating natural language explanations of knowledge graph rules.
\newblock \emph{arXiv preprint arXiv:2508.10971}, 2025.

\bibitem[Wang et~al.(2022)Wang, Deng, Sun, and Meng]{wang2022perplexity}
Yequan Wang, Jiawen Deng, Aixin Sun, and Xuying Meng.
\newblock Perplexity from plm is unreliable for evaluating text quality.
\newblock \emph{arXiv preprint arXiv:2210.05892}, 2022.

\bibitem[Hashimoto et~al.(2019)Hashimoto, Zhang, and Liang]{hashimoto2019unifying}
Tatsunori~B. Hashimoto, Hugh Zhang, and Percy Liang.
\newblock Unifying human and statistical evaluation for natural language generation.
\newblock In \emph{Proceedings of the 2019 Conference of the North {A}merican Chapter of the Association for Computational Linguistics: Human Language Technologies, Volume 1 (Long and Short Papers)}, pages 1689--1701, Minneapolis, Minnesota, 2019. Association for Computational Linguistics.
\newblock \doi{10.18653/v1/N19-1169}.

\bibitem[Chaganty et~al.(2018)Chaganty, Mussman, and Liang]{chaganty2018price}
Arun~Tejasvi Chaganty, Stephen Mussman, and Percy Liang.
\newblock The price of debiasing automatic metrics in natural language evaluation.
\newblock \emph{arXiv preprint arXiv:1807.02202}, 2018.

\bibitem[Veli{\v{c}}kovi{\'c} et~al.(2026)Veli{\v{c}}kovi{\'c}, Barbero, Perivolaropoulos, Osindero, and Pascanu]{velivckovic2026perplexity}
Petar Veli{\v{c}}kovi{\'c}, Federico Barbero, Christos Perivolaropoulos, Simon Osindero, and Razvan Pascanu.
\newblock Perplexity cannot always tell right from wrong.
\newblock \emph{arXiv preprint arXiv:2601.22950}, 2026.

\bibitem[Fang et~al.(2024)Fang, Wang, Liu, Zhang, Jegelka, Gao, Ding, and Wang]{fang2024wrong}
Lizhe Fang, Yifei Wang, Zhaoyang Liu, Chenheng Zhang, Stefanie Jegelka, Jinyang Gao, Bolin Ding, and Yisen Wang.
\newblock What is wrong with perplexity for long-context language modeling?
\newblock \emph{arXiv preprint arXiv:2410.23771}, 2024.

\bibitem[Papineni et~al.(2002)Papineni, Roukos, Ward, and Zhu]{papineni2002bleu}
Kishore Papineni, Salim Roukos, Todd Ward, and Wei-Jing Zhu.
\newblock Bleu: a method for automatic evaluation of machine translation.
\newblock In \emph{Proceedings of the 40th annual meeting of the Association for Computational Linguistics}, pages 311--318, 2002.

\bibitem[Banerjee and Lavie(2005)]{banerjee2005meteor}
Satanjeev Banerjee and Alon Lavie.
\newblock Meteor: An automatic metric for mt evaluation with improved correlation with human judgments.
\newblock In \emph{Proceedings of the acl workshop on intrinsic and extrinsic evaluation measures for machine translation and/or summarization}, pages 65--72, 2005.

\bibitem[Lin(2004)]{lin2004rouge}
Chin-Yew Lin.
\newblock Rouge: A package for automatic evaluation of summaries.
\newblock In \emph{Text summarization branches out}, pages 74--81, 2004.

\bibitem[Sellam et~al.(2020)Sellam, Das, and Parikh]{sellam2020bleurt}
Thibault Sellam, Dipanjan Das, and Ankur Parikh.
\newblock Bleurt: Learning robust metrics for text generation.
\newblock In \emph{Proceedings of the 58th annual meeting of the association for computational linguistics}, pages 7881--7892, 2020.

\bibitem[Meister and Cotterell(2021)]{meister2021language}
Clara Meister and Ryan Cotterell.
\newblock Language model evaluation beyond perplexity.
\newblock In \emph{Proceedings of the 59th Annual Meeting of the Association for Computational Linguistics and the 11th International Joint Conference on Natural Language Processing (Volume 1: Long Papers)}, pages 5328--5339. Association for Computational Linguistics, 2021.
\newblock \doi{10.18653/v1/2021.acl-long.414}.

\bibitem[Devlin et~al.(2019)Devlin, Chang, Lee, and Toutanova]{devlin2019bert}
Jacob Devlin, Ming-Wei Chang, Kenton Lee, and Kristina Toutanova.
\newblock Bert: Pre-training of deep bidirectional transformers for language understanding.
\newblock In \emph{Proceedings of the 2019 conference of the North American chapter of the association for computational linguistics: human language technologies, volume 1 (long and short papers)}, pages 4171--4186, 2019.

\bibitem[Tekkesinoglu and Kunze(2024)]{tekkesinoglu2024feature}
Sule Tekkesinoglu and Lars Kunze.
\newblock From feature importance to natural language explanations using llms with rag.
\newblock \emph{arXiv preprint arXiv:2407.20990}, 2024.

\bibitem[Graesser et~al.(2004)Graesser, McNamara, Louwerse, and Cai]{graesser2004coh}
Arthur~C Graesser, Danielle~S McNamara, Max~M Louwerse, and Zhiqiang Cai.
\newblock Coh-metrix: Analysis of text on cohesion and language.
\newblock \emph{Behavior research methods, instruments, \& computers}, 36\penalty0 (2):\penalty0 193--202, 2004.

\bibitem[Sap et~al.(2020)Sap, Horvitz, Choi, Smith, and Pennebaker]{sap2020recollection}
Maarten Sap, Eric Horvitz, Yejin Choi, Noah~A Smith, and James Pennebaker.
\newblock Recollection versus imagination: Exploring human memory and cognition via neural language models.
\newblock In \emph{Proceedings of the 58th annual meeting of the association for computational linguistics}, pages 1970--1978, 2020.

\bibitem[Graesser and McNamara(2011)]{graesser_computationalanalyses}
Arthur~C. Graesser and Danielle~S. McNamara.
\newblock Computational analyses of multilevel discourse comprehension.
\newblock \emph{Topics in Cognitive Science}, 3\penalty0 (2):\penalty0 371--398, 2011.
\newblock \doi{https://doi.org/10.1111/j.1756-8765.2010.01081.x}.
\newblock URL \url{https://onlinelibrary.wiley.com/doi/abs/10.1111/j.1756-8765.2010.01081.x}.

\bibitem[Zhou et~al.(2021)Zhou, Gandomi, Chen, and Holzinger]{electronics10050593}
Jianlong Zhou, Amir~H. Gandomi, Fang Chen, and Andreas Holzinger.
\newblock Evaluating the quality of machine learning explanations: A survey on methods and metrics.
\newblock \emph{Electronics}, 10\penalty0 (5), 2021.
\newblock ISSN 2079-9292.
\newblock \doi{10.3390/electronics10050593}.
\newblock URL \url{https://www.mdpi.com/2079-9292/10/5/593}.

\bibitem[Salmon(1984)]{salmon1984scientific}
Wesley~C Salmon.
\newblock Scientific explanation and the causal structure of the world.
\newblock 1984.

\bibitem[Salmon(1989)]{salmon1989four}
Wesley~C Salmon.
\newblock Four decades of scientific explanation.[part 3] the second decade (1958-67): Manifest destiny--expansion and conflict.
\newblock 1989.

\bibitem[Murray(1997)]{murray1997connectives}
John~D Murray.
\newblock Connectives and narrative text: The role of continuity.
\newblock \emph{Memory \& Cognition}, 25\penalty0 (2):\penalty0 227--236, 1997.

\bibitem[Graesser et~al.(2011)Graesser, McNamara, and Kulikowich]{graesser2011coh}
Arthur~C Graesser, Danielle~S McNamara, and Jonna~M Kulikowich.
\newblock Coh-metrix: Providing multilevel analyses of text characteristics.
\newblock \emph{Educational researcher}, 40\penalty0 (5):\penalty0 223--234, 2011.

\bibitem[Galbraith(1995)]{galbraith1995deictic}
Mary Galbraith.
\newblock Deictic shift theory and the poetics of involvement in narrative.
\newblock In \emph{Deixis in narrative}, pages 19--59. Psychology Press, 1995.

\bibitem[O'Brien et~al.(1998)O'Brien, Rizzella, Albrecht, and Halleran]{o1998updating}
Edward~J O'Brien, Michelle~L Rizzella, Jason~E Albrecht, and Jennifer~G Halleran.
\newblock Updating a situation model: a memory-based text processing view.
\newblock \emph{Journal of Experimental Psychology: Learning, Memory, and Cognition}, 24\penalty0 (5):\penalty0 1200, 1998.

\bibitem[Simon(2000)]{simon2000discovering}
Herbert~A Simon.
\newblock Discovering explanations.
\newblock \emph{Explanation and cognition}, pages 21--59, 2000.

\bibitem[Hempel and Oppenheim(1948)]{hempel1948studies}
Carl~G Hempel and Paul Oppenheim.
\newblock Studies in the logic of explanation.
\newblock \emph{Philosophy of science}, 15\penalty0 (2):\penalty0 135--175, 1948.

\bibitem[Kellogg et~al.(1991)]{kellogg1991relative}
Ronald~T Kellogg et~al.
\newblock The relative ease of writing narrative text.
\newblock 1991.

\bibitem[Saenz and Fuchs(2002)]{saenz2002examining}
Laura~M Saenz and Lynn~S Fuchs.
\newblock Examining the reading difficulty of secondary students with learning disabilities: Expository versus narrative text.
\newblock \emph{Remedial and Special Education}, 23\penalty0 (1):\penalty0 31--41, 2002.

\bibitem[Weaver~III and Kintsch(1991)]{weaver1991expository}
Charles~A Weaver~III and Walter Kintsch.
\newblock Expository text.
\newblock 1991.

\bibitem[Shannon(1951)]{shannon1951prediction}
Claude~E Shannon.
\newblock Prediction and entropy of printed english.
\newblock \emph{Bell system technical journal}, 30\penalty0 (1):\penalty0 50--64, 1951.

\bibitem[Holtzman et~al.(2019)Holtzman, Buys, Du, Forbes, and Choi]{holtzman2019curious}
Ari Holtzman, Jan Buys, Li~Du, Maxwell Forbes, and Yejin Choi.
\newblock The curious case of neural text degeneration.
\newblock \emph{arXiv preprint arXiv:1904.09751}, 2019.

\bibitem[Das et~al.(2018)Das, Scheffler, Bourgonje, and Stede]{das2018constructing}
Debopam Das, Tatjana Scheffler, Peter Bourgonje, and Manfred Stede.
\newblock Constructing a lexicon of english discourse connectives.
\newblock In \emph{Proceedings of the 19th Annual SIGdial Meeting on Discourse and Dialogue}, pages 360--365, 2018.

\bibitem[Valentino and Freitas(2022)]{valentino2022scientific}
Marco Valentino and Andr{\'e} Freitas.
\newblock Scientific explanation and natural language: A unified epistemological-linguistic perspective for explainable ai.
\newblock \emph{arXiv preprint arXiv:2205.01809}, 2022.

\bibitem[Valentino and Freitas(2024)]{valentino2024nature}
Marco Valentino and Andr{\'e} Freitas.
\newblock On the nature of explanation: An epistemological-linguistic perspective for explanation-based natural language inference.
\newblock \emph{Philosophy \& Technology}, 37\penalty0 (3):\penalty0 88, 2024.

\bibitem[Kitcher(1981)]{kitcher1981}
Philip Kitcher.
\newblock Explanatory unification.
\newblock \emph{Philosophy of Science}, 48\penalty0 (4):\penalty0 507--531, 1981.
\newblock ISSN 00318248, 1539767X.
\newblock URL \url{http://www.jstor.org/stable/186834}.

\bibitem[Hempel et~al.(1965)]{hempel1965aspects}
Carl~G Hempel et~al.
\newblock \emph{Aspects of scientific explanation}, volume 965.
\newblock Free press New York, 1965.

\bibitem[Prasad et~al.(2019)Prasad, Webber, Lee, and Joshi]{prasad2019penn}
Rashmi Prasad, Bonnie Webber, Alan Lee, and Aravind Joshi.
\newblock Penn discourse treebank version 3.0.
\newblock \emph{LDC2019T05}, 2019.

\bibitem[Grave et~al.(2016)Grave, Joulin, and Usunier]{grave2016improving}
Edouard Grave, Armand Joulin, and Nicolas Usunier.
\newblock Improving neural language models with a continuous cache.
\newblock \emph{arXiv preprint arXiv:1612.04426}, 2016.

\bibitem[Lee et~al.(2022)Lee, Ping, Xu, Patwary, Fung, Shoeybi, and Catanzaro]{lee2022_factualityenhanced}
Nayeon Lee, Wei Ping, Peng Xu, Mostofa Patwary, Pascale~N Fung, Mohammad Shoeybi, and Bryan Catanzaro.
\newblock Factuality enhanced language models for open-ended text generation.
\newblock In \emph{Advances in Neural Information Processing Systems}, volume~35, pages 34586--34599. Curran Associates, Inc., 2022.
\newblock URL \url{https://proceedings.neurips.cc/paper_files/paper/2022/file/df438caa36714f69277daa92d608dd63-Paper-Conference.pdf}.

\bibitem[Li et~al.(2016)Li, Galley, Brockett, Gao, and Dolan]{li2016diversity}
Jiwei Li, Michel Galley, Chris Brockett, Jianfeng Gao, and William~B Dolan.
\newblock A diversity-promoting objective function for neural conversation models.
\newblock In \emph{Proceedings of the 2016 conference of the North American chapter of the association for computational linguistics: human language technologies}, pages 110--119, 2016.

\bibitem[Bao and Zeng(2024)]{bao2024understanding}
Aorigele Bao and Yi~Zeng.
\newblock Understanding the dilemma of explainable artificial intelligence: a proposal for a ritual dialog framework.
\newblock \emph{Humanities and Social Sciences Communications}, 11\penalty0 (1):\penalty0 1--9, 2024.

\bibitem[Breiman(2001)]{breiman2001random}
Leo Breiman.
\newblock Random forests.
\newblock \emph{Machine learning}, 45\penalty0 (1):\penalty0 5--32, 2001.

\bibitem[Domnich et~al.(2025)Domnich, V{\"a}lja, Veski, Magnifico, Tulver, Barbu, and Vicente]{domnich2025towards}
Marharyta Domnich, Julius V{\"a}lja, Rasmus~Moorits Veski, Giacomo Magnifico, Kadi Tulver, Eduard Barbu, and Raul Vicente.
\newblock Towards unifying evaluation of counterfactual explanations: Leveraging large language models for human-centric assessments.
\newblock In \emph{Proceedings of the AAAI Conference on Artificial Intelligence}, volume~39, pages 16308--16316, 2025.

\bibitem[Achiam et~al.(2023)Achiam, Adler, Agarwal, Ahmad, Akkaya, Aleman, Almeida, Altenschmidt, Altman, Anadkat, et~al.]{achiam2023gpt}
Josh Achiam, Steven Adler, Sandhini Agarwal, Lama Ahmad, Ilge Akkaya, Florencia~Leoni Aleman, Diogo Almeida, Janko Altenschmidt, Sam Altman, Shyamal Anadkat, et~al.
\newblock Gpt-4 technical report.
\newblock \emph{arXiv preprint arXiv:2303.08774}, 2023.

\bibitem[Grattafiori et~al.(2024)Grattafiori, Dubey, Jauhri, Pandey, Kadian, Al-Dahle, Letman, Mathur, Schelten, Vaughan, et~al.]{grattafiori2024llama}
Aaron Grattafiori, Abhimanyu Dubey, Abhinav Jauhri, Abhinav Pandey, Abhishek Kadian, Ahmad Al-Dahle, Aiesha Letman, Akhil Mathur, Alan Schelten, Alex Vaughan, et~al.
\newblock The llama 3 herd of models.
\newblock \emph{arXiv preprint arXiv:2407.21783}, 2024.

\bibitem[Dem{\v{s}}ar(2006)]{demvsar2006statistical}
Janez Dem{\v{s}}ar.
\newblock Statistical comparisons of classifiers over multiple data sets.
\newblock \emph{Journal of Machine learning research}, 7\penalty0 (Jan):\penalty0 1--30, 2006.

\bibitem[Tikhonov et~al.(2021)Tikhonov, Samenko, and Yamshchikov]{tikhonov2021storydb}
Alexey Tikhonov, Igor Samenko, and Ivan~P. Yamshchikov.
\newblock {S}tory{DB}: Broad multi-language narrative dataset.
\newblock In Yang Gao, Steffen Eger, Wei Zhao, Piyawat Lertvittayakumjorn, and Marina Fomicheva, editors, \emph{Proceedings of the 2nd Workshop on Evaluation and Comparison of NLP Systems}, pages 32--39, Punta Cana, Dominican Republic, November 2021. Association for Computational Linguistics.
\newblock \doi{10.18653/v1/2021.eval4nlp-1.4}.
\newblock URL \url{https://aclanthology.org/2021.eval4nlp-1.4/}.

\end{thebibliography}

\clearpage
\appendix
\section{Cause-Effect Markers Lexicon}
\label{app:cause_effect_lex}

In the following, we present the final list of the Cause-Effect markers that are used to calculate the Cause-Effect Ratio (CER). The markers are distilled from Lexicon of English Discourse Connectives~\cite{das2018constructing} and include: accordingly, as a result, as a result of, because, because of, consequently, due to, given, given that, hence, in response to, in this way, in turn, since, so, so that, thereby, therefore, thus.

\section{Prompt examples for LLM-generated explanations}
\subsection{XAIstories}
\label{app:xaistories_prompt}
\vspace{-1em}
Below, we present the prompt used to generate explanations using XAIstories method~\cite{tellmeastory}.
\begin{tcolorbox}[tightbox, title=\textbf{XAIstories Prompt}]
\setlength{\parskip}{5pt}
**Format-related rules**:
\begin{enumerate}
\item Start the explanation immediately.
\item Limit the entire answer to exactly \{sentence\_limit\} sentences.
\item Only mention the top \{num\_feat\} most important features in the narrative.
\item Do not use tables or lists, or simply rattle through the features and/or nodes one by one. The goal is to have a narrative/story.
\end{enumerate}

**Content related rules**:
\begin{enumerate} 
\item Be clear about what the model actually predicted for the \{target\_instance\}.
\item Discuss how the features contributed to final prediction. Make sure to clearly establish this the first time you refer to a feature. 
\item Consider the feature importance, feature values, and averages when referencing their relative importance.
\item Begin the discussion of features by presenting those with the highest absolute feature importance values first. The reader should be able to tell what the order of importance of the features is based on their feature importance value.
\item Provide a suggestion or interpretation as to why a feature contributed in a certain direction. Try to introduce external knowledge that you might have.
\item If there is no simple explanation for the effect of a feature, consider the context of other features in the interpretation.
\item Do not use the feature importance numeric values in your answer.
\item You can use the feature values themselves in the explanation, as long as they are not categorical variables. If they are an categorical variable, refer to the semantic meaning of the category from the \{feature\_desc\}.
\item Do not refer to the average for every single feature; reserve it for features where it truly clarifies the explanation.
\item When you refer to an instance, keep in mind that the target instance is a \{target\_instance\}.
\item Tell a clear and engaging story, including details from both feature values and node connections, to make the explanation more relatable and interesting.
\item Use clear and simple language that a general audience can understand, avoiding overly technical jargon or explaining any necessary technical terms in plain language.
\end{enumerate}
\end{tcolorbox}

\clearpage
\subsection{Explingo}
\label{app:explingo_prompt}

Below, we present the prompts used to generate explanations using Explingo method~\cite{zytek2024explingo}.

\begin{tcolorbox}[tightbox, title=\textbf{Explingo Prompt Example}]
\setlength{\parskip}{5pt}
You are helping users understand an ML model’s prediction. Given an explanation and information about the model, convert
the explanation into a human-readable narrative.

Follow the following format.
Context: What the ML model predicts
Explanation: Explanation of an ML model’s prediction
Explanation format: Format the explanation is given in
Narrative: Human-readable narrative version of the explanation

Explanation: (Living area square feet, 2090.00, 16382.07), (Second floor square feet, 983.00, 16216.99)
Explanation format: (feature\_name, feature\_value, SHAP feature contribution)
Context: The ML model predicts house prices

Please provide the output field Narrative. Do so immediately, without additional content before or after, and precisely as the
format above shows. Begin with only the field Narrative.
\end{tcolorbox}

\section{Diversity-adjusted Context Progression Rate fitting}
\label{app:fitting_table}

In the Table~\ref{tab:fitted_dcpr}, we present the results of the DCPR metrics with the fitted $r$ parameter, and the corresponding goodness-of-fit results for $R^2$ ($\uparrow$) and $RMSE$ ($\downarrow$).

\begin{table}[h!]
\centering
\caption{DCPR results with fitted $r$ across datasets and methods.}
\label{tab:fitted_dcpr}
\begin{tabular}{llrrr}
\toprule
Dataset & Method & DCPR $\downarrow$ & $R^2$ $\uparrow$ & RMSE $\downarrow$ \\
\midrule
\multirow{5}{*}{Compas} 
 & Templated Explanation & 0.59 & \textbf{1.00} & 0.39 \\
 & Explingo Zero Shot & 0.62 & \textbf{1.00} & 0.32 \\
 & Explingo Narratives & 0.50 & 0.99 & 0.44 \\
 & XAIstories & 0.44 & \textbf{1.00} & \textbf{0.16} \\
 & XAIstories Narratives & \textbf{0.42} & 0.99 & 0.39 \\
\midrule
\multirow{5}{*}{Diabetes} 
 & Templated Explanation & 0.52 & \textbf{1.00} & 0.50 \\
 & Explingo Zero Shot & 0.39 & \textbf{1.00} & 0.24 \\
 & Explingo Narratives & 0.40 & 0.98 & 1.51 \\
 & XAIstories & \textbf{0.28} & \textbf{1.00} & 0.40 \\
 & XAIstories Narratives & 0.40 & \textbf{1.00} & \textbf{0.20} \\
\midrule
\multirow{5}{*}{Fifa}
 & Templated Explanation & 1.00 & \textbf{1.00} & 0.86 \\
 & Explingo Zero Shot & 0.55 & \textbf{1.00} & \textbf{0.14} \\
 & Explingo Narratives & 0.67 & \textbf{1.00} & 0.19 \\
 & XAIstories & 0.42 & 0.99 & 0.54 \\
 & XAIstories Narratives & 0.63 & \textbf{1.00} & 0.30 \\
\midrule
\multirow{5}{*}{German Credit}
 & Templated Explanation & \textbf{0.38} & \textbf{1.00} & 0.91 \\
 & Explingo Zero Shot & 0.45 & 0.99 & 0.80 \\
 & Explingo Narratives & 0.44 & \textbf{1.00} & 0.38 \\
 & XAIstories & 0.41 & \textbf{1.00} & \textbf{0.30} \\
 & XAIstories Narratives & 0.44 & \textbf{1.00} & 0.42 \\
\midrule
\multirow{5}{*}{Stroke}
 & Templated Explanation & 0.58 & \textbf{1.00} & 0.33 \\
 & Explingo Zero Shot & 0.40 & 0.99 & 1.41 \\
 & Explingo Narratives & 0.45 & 0.99 & 0.63 \\
 & XAIstories & \textbf{0.35} & 0.99 & 0.38 \\
 & XAIstories Narratives & 0.48 & \textbf{1.00} & \textbf{0.16} \\
\midrule
\multirow{5}{*}{Student}
 & Templated Explanation & 0.81 & \textbf{1.00} & 1.71 \\
 & Explingo Zero Shot & 0.60 & 0.99 & 0.89 \\
 & Explingo Narratives & 0.65 & 0.98 & 0.99 \\
 & XAIstories & 0.40 & \textbf{1.00} & \textbf{0.34} \\
 & XAIstories Narratives & 0.54 & \textbf{1.00} & 0.36 \\
\bottomrule
\end{tabular}
\end{table}

\section{Additional Analysis}
\label{app:charts}

\begin{figure}[h!] 
    \centering 
    \begin{adjustbox}{width=1\linewidth,center} 
        \includegraphics{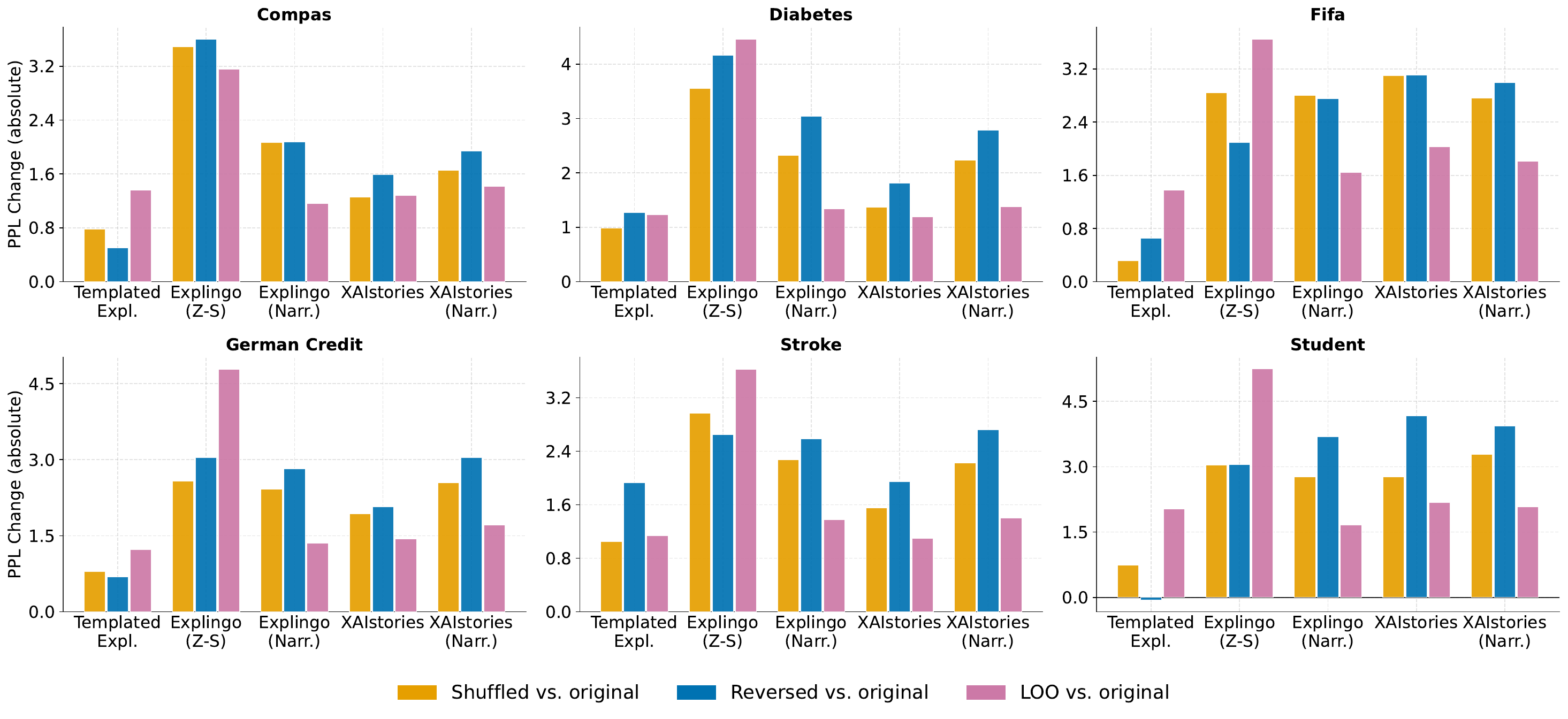} 
    \end{adjustbox} 
    \caption{Nominal change in perplexity (PPL) relative to the PPL of the original sentence order across six tabular datasets for five explanation methods. Bars report increases in PPL under the shuffled, reversed, and leave‑one‑out sentence reordering.} 
    \label{fig:ppl_relative_increase} 
\end{figure}

To complete the \emph{Continuous Structure Rate}, which measures sensitivity to random sentence reordering, we also measure perplexity changes under two additional perturbations to sentence ordering: sentence reversal and leave-one-out deletion. These analyses test whether an explanation behaves like an unordered list of statements or shows order-dependent structure. Results appear in Figure~\ref{fig:ppl_relative_increase}. We evaluate the robustness of each explanation method by measuring the nominal difference in perplexity between the perturbed and original texts. Templated Explanations exhibit a considerably lower baseline perplexity than the narrative approaches. Expressing the changes as relative percentages artificially inflates the impact of perturbations on the templated baseline due to the small original values of the PPL. Analysing the nominal difference isolates the increase in language model surprise and allows for a fair comparison across methods with varying initial perplexity levels. The results demonstrate that structural perturbations cause larger increases in perplexity for the narrative explanations than for the templated baselines. This result aligns with the expectation that complex narrative structures are more sensitive to ordering and content omission than simple templates. Reordering the Templated Explanations on the Student dataset yields even lower perplexity under sentence reversal, which indicates that the text does not rely on a consistent sentence order and behaves more like a rearrangeable feature list than an order-dependent narrative.

\begin{figure}[t!] 
  \centering
  \begin{adjustbox}{width=1\linewidth,center}
    \includegraphics{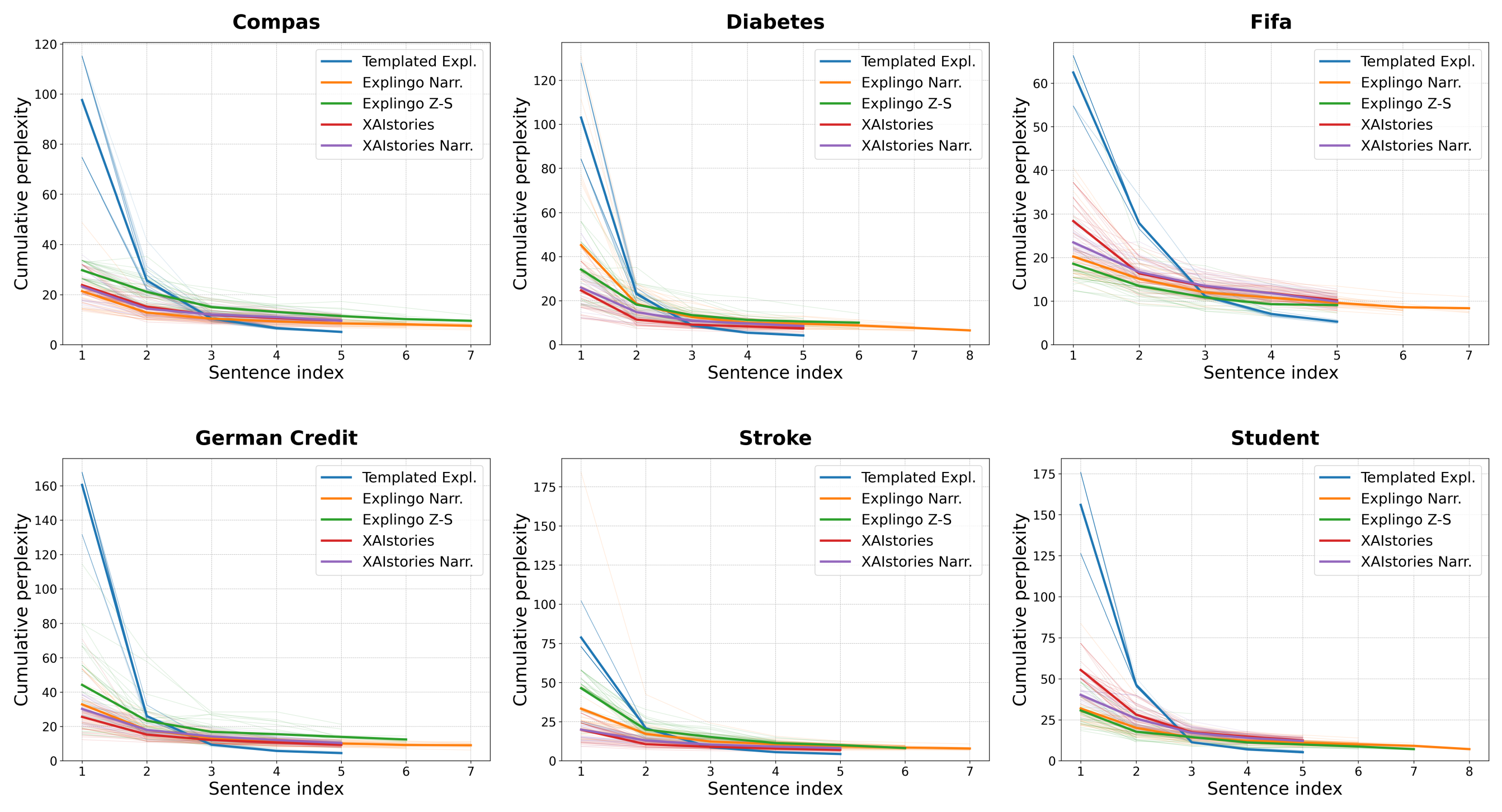}
    \end{adjustbox}
    \caption{Cumulative sentence perplexity trajectories of templated and LLM-generated explanatory methods across datasets. The consistently lower values for the Templated Explanation reflect structural repetition rather than narrative fluency.}
    \label{fig:cumulaive_perplexity_frameworks}
\end{figure}

\begin{figure}[!h]
    \centering 
    \begin{adjustbox}{width=1\linewidth,center} 
    \includegraphics{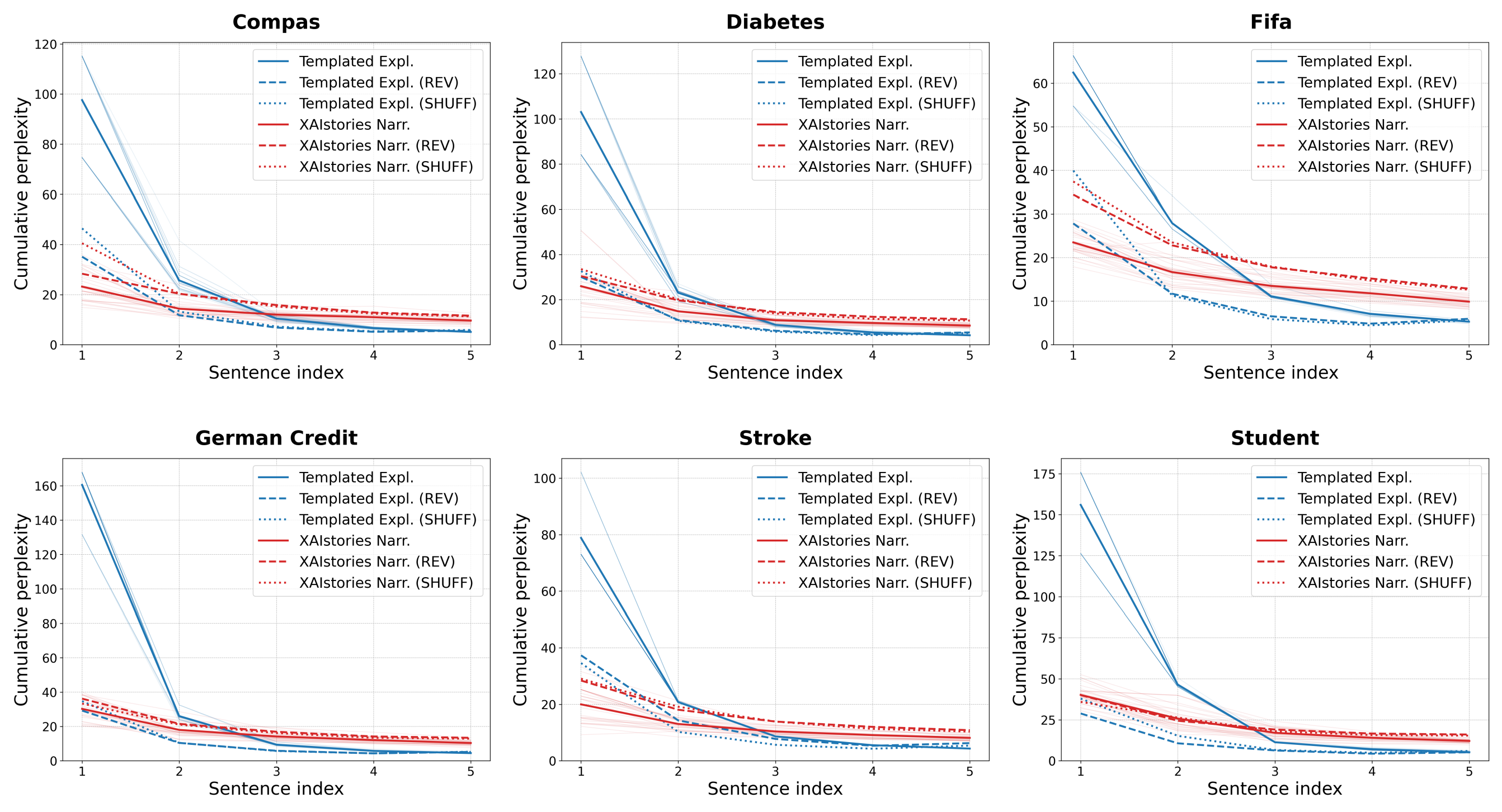} 
    \end{adjustbox} 
    \caption{Changes in cumulative perplexity after sentence ordering perturbations of Templated Explanations and XAIstories Narratives methods.} 
    \label{fig:cumulaive_perplexity_sentence_order} 
\end{figure}

Further, in the Figure~\ref{fig:cumulaive_perplexity_frameworks} and Figure~\ref{fig:cumulaive_perplexity_sentence_order}, we present the cumulative perplexity trajectories for the analysed methods across datasets. TalkToModel and the original Explingo methods are omitted as their single-sentence outputs do not permit sentence-level cumulative analysis. Among the remaining approaches, Templated Explanation attains the lowest final cumulative perplexity, which is consistent with repeated phrasing in a fixed template rather than with discourse progression~\cite{wang2022perplexity} which is shown in Figure~\ref{fig:cumulaive_perplexity_frameworks}.

Figure~\ref{fig:cumulaive_perplexity_sentence_order} shows the comparison of the cumulative perplexity for Templated Explanation and XAIstories Narratives under sentence shuffling and reversal. For XAIstories Narratives, both perturbations increase cumulative perplexity relative to the original order, which aligns with order-dependent coherence. For the templated method, cumulative perplexity changes little under perturbation and in some cases decreases early in the sequence. This behaviour is consistent with weak sequential dependence and suggests that the text is closer to a reorderable feature list than to an order-sensitive narrative. Moreover, both Figure~\ref{fig:cumulaive_perplexity_frameworks} 
and Figure~\ref{fig:cumulaive_perplexity_sentence_order} provide further motivation for fitting the exponential function to the cumulative perplexities, as done in Equation~\eqref{eq:exp_fit}, whose decay rate $r$ we subsequently use to compute our metrics.

\begin{figure}[!h] 
\centering 
\begin{adjustbox}{width=1\linewidth,center} 
\includegraphics{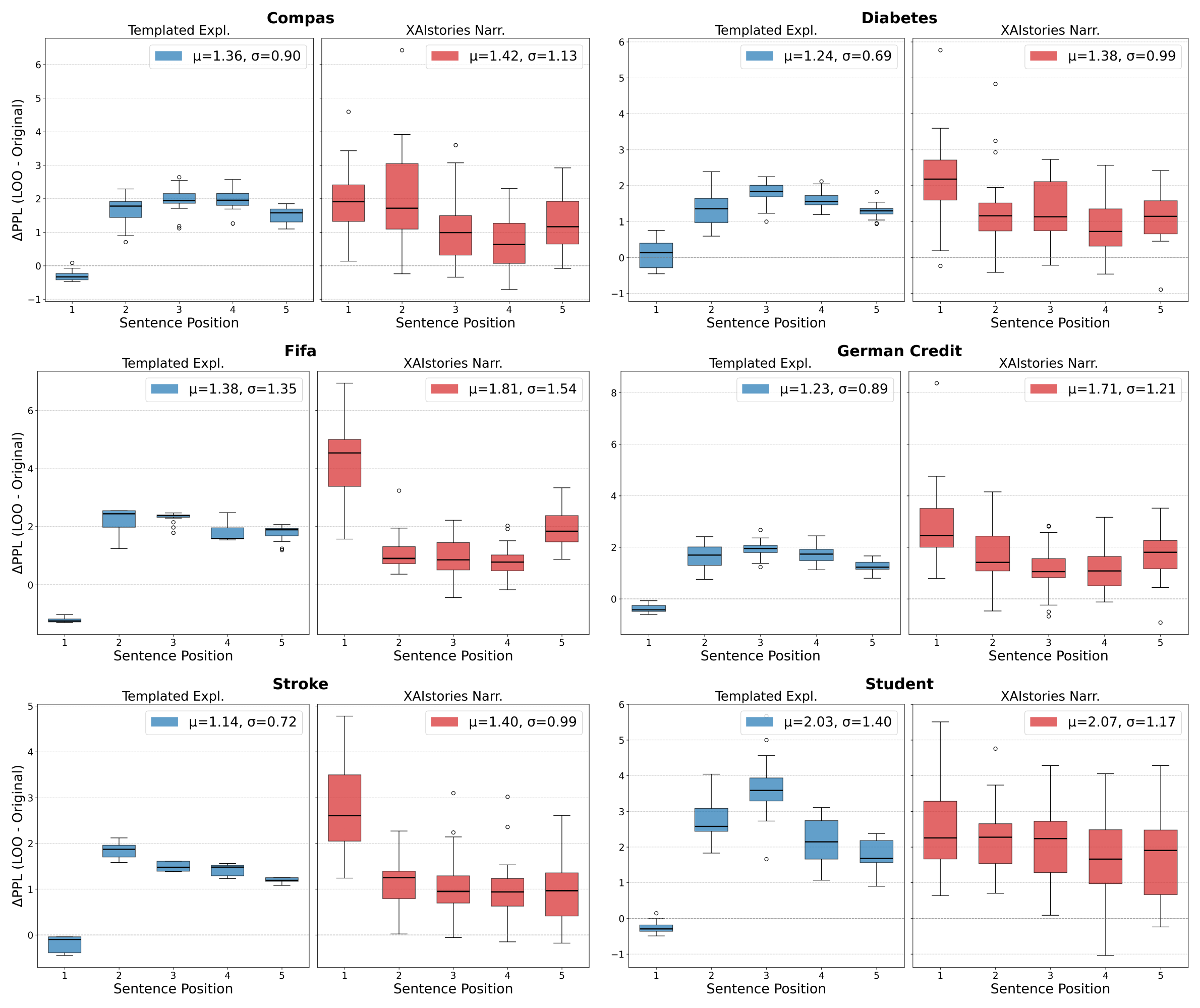} 
\end{adjustbox} 
\caption{Changes in the in the perplexity distribution at particular sentence position of the explanation after a single-sentence ablation (leave-one-out).}
\label{fig:cumulaive_perplexity_distribution} 
\end{figure}

\clearpage
Finally, in Figure~\ref{fig:cumulaive_perplexity_distribution} we show distributional changes in perplexity under single-sentence ablation (leave-one-out approach). Across all datasets, XAIstories Narratives show larger mean changes and higher variance than Templated Explanation. This pattern is consistent with stronger inter-sentence dependence in narrative explanations and weaker dependence in templated text, which remains comparatively insensitive to removing individual sentences.

\end{document}